%% file: main.tex
\documentclass[10pt,twocolumn,letterpaper]{article}

\usepackage[pagenumbers]{cvpr} 
\usepackage{graphicx}
\usepackage{caption}
\usepackage{subcaption}
\usepackage{array}
\usepackage{epigraph}
\usepackage{multirow}
\usepackage{booktabs}
\usepackage{pifont}
\usepackage{float}
\usepackage{url}            

\usepackage{url}            
\newcommand\codeurl[1]{{{\color{blue}{\url{#1}}}}}
\input{preamble}

\definecolor{cvprblue}{rgb}{0.21,0.49,0.74}
\usepackage[pagebackref,breaklinks,colorlinks,citecolor=cvprblue]{hyperref}

\newlength\savewidth\newcommand\shline{\noalign{\global\savewidth\arrayrulewidth
  \global\arrayrulewidth 1pt}\hline\noalign{\global\arrayrulewidth\savewidth}}
\newcommand{\tablestyle}[2]{\setlength{\tabcolsep}{#1}\renewcommand{\arraystretch}{#2}\centering\footnotesize}

\newcommand{\dalle}{DALL$\cdot$E3\xspace}

\newcommand\rgt{\aftergroup\mathclose\aftergroup{\aftergroup}\right}

\makeatletter
\newcommand{\removelatexerror}{\let\@latex@error\@gobble}
\makeatother

\usepackage{algorithm}
\usepackage{algorithmic}

\title{DesignEdit: Multi-Layered Latent Decomposition and Fusion for\\ Unified \& Accurate Image Editing}

\author{
{\normalsize Yueru Jia$^{1}$ \quad Yuhui Yuan$^{1,2,3}$ \quad Aosong Cheng
 \quad Chuke Wang \quad Ji Li \quad Huizhu Jia \quad Shanghang Zhang$^{3}$}\\[0mm]
 {\small$^1$joint core contribution \qquad $^2$project lead \qquad $^3$corresponding author}\\[1mm]
\normalsize{Microsoft Research Asia\quad\quad Peking University}\\
{\footnotesize\codeurl{{https://design-edit.github.io/}}}\vspace{-4mm}}

\begin{document}

\twocolumn[
{%
\renewcommand\twocolumn[1][]{#1}
\maketitle
\begin{center}
\centering
\begin{minipage}[t]{\linewidth}
\includegraphics[width=\textwidth]{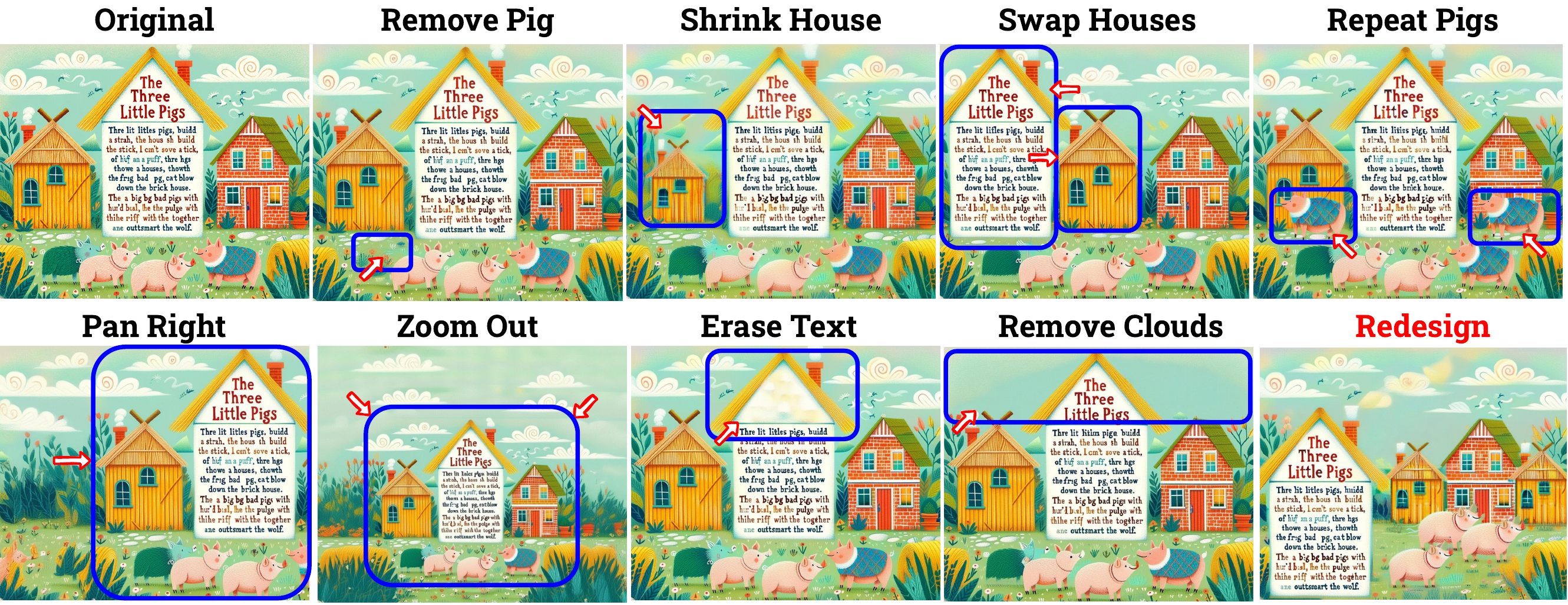}
{\captionsetup{hypcap=false}  
\captionof{figure}{\footnotesize{\textbf{Examples of visual design image editing.} Our approach facilitates a range of image editing operations with a training-free and unified framework to achieve accurate spatial-aware editing of the design image. Our approach is able to manipulate different objects simultaneously, as well as implement various operations at the same time. All results are produced using one diffusion denoising process.}}
\label{fig:teaser}}
\end{minipage}
\end{center}
}]
\input{sec/0_abstract}
\input{sec/1_intro}
\input{sec/2_relate}
\input{sec/3_approach}
\input{sec/4_experiment}
\input{sec/5_conclusion}
{
    \small
    \bibliographystyle{ieeenat_fullname}
    \bibliography{main}
}

\end{document}

%% file: preamble.tex
\usepackage[dvipsnames]{xcolor}


%% file: sec/0_abstract.tex
\begin{abstract}
Recently, how to achieve precise image editing has attracted increasing attention, especially given the remarkable success of text-to-image generation models. To unify various spatial-aware image editing abilities into one framework, we adopt the concept of layers from the design domain to manipulate objects flexibly with various operations. The key insight is to transform the spatial-aware image editing task into a combination of two sub-tasks: \emph{multi-layered latent decomposition} and \emph{multi-layered latent fusion}. First, we segment the latent representations of the source images into multiple layers, which include several object layers and one incomplete background layer that necessitates reliable inpainting. To avoid extra tuning, we further explore the inner inpainting ability within the self-attention mechanism. We introduce a key-masking self-attention scheme that can propagate the surrounding context information into the masked region while mitigating its impact on the regions outside the mask. Second, we propose an instruction-guided latent fusion that pastes the multi-layered latent representations onto a canvas latent. We also introduce an artifact suppression scheme in the latent space to enhance the inpainting quality. Due to the inherent modular advantages of such multi-layered representations, we can achieve accurate image editing, and we demonstrate that our approach consistently surpasses the latest spatial editing methods, including Self-Guidance and DiffEditor. Last, we show that our approach is a unified framework that supports various accurate image editing tasks on more than six different editing tasks.
\end{abstract}

%% file: sec/1_intro.tex
\section{Introduction}
\label{sec:intro}

\begin{figure*}[!htbp] 
\centering
\includegraphics[width=\textwidth]{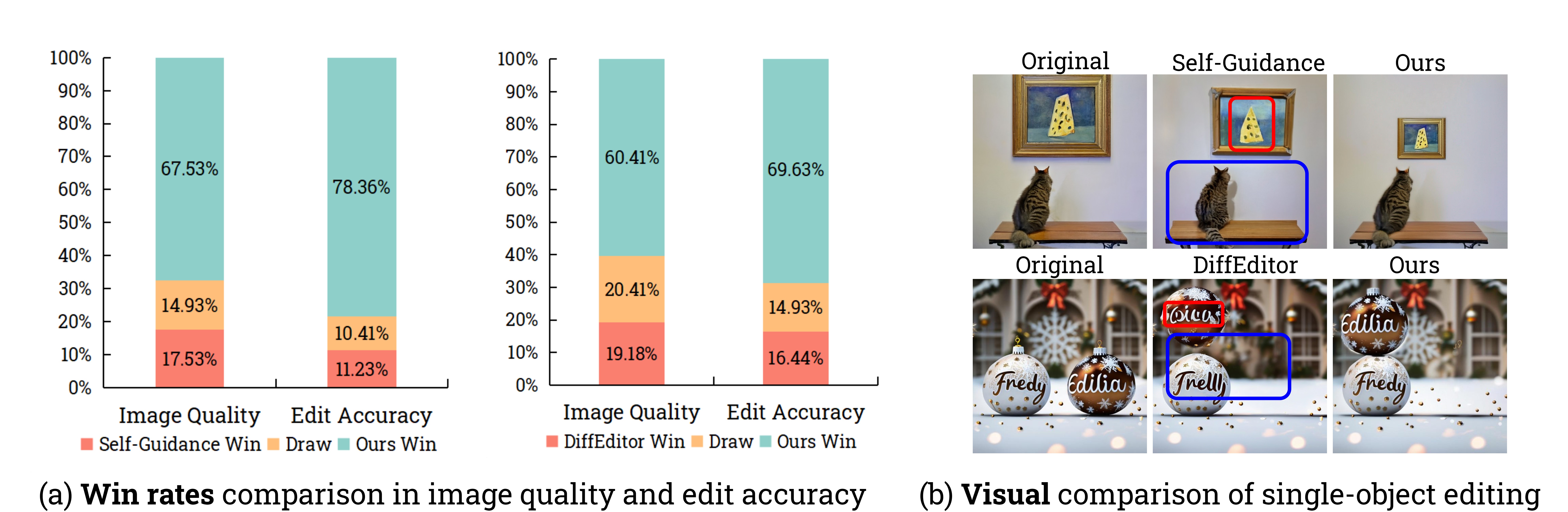}
\caption{\footnotesize
\textbf{Comparison between our method against Self-Guidance and DiffEditor.} We report the win-rate comparison across image quality and edit accuracy in (a).  For each comparison, we select 10 examples with multiple operations like movement and resizing. Users were asked to vote from two aspects, image quality and edit accuracy. The ``Draw'' option represents equal effect. We collect answers from 73 users, with a total of 1460 votes for each metric.}
\label{fig:user_study}
\end{figure*}

Despite the great achievements in image generation by training large-scale text-to-image diffusion models~\cite{nichol2022glide, ramesh2022hierarchical, saharia2022photorealistic, ruiz2023dreambooth, gu2022vector, kawar2023imagic}, as demonstrated by recent seminal research including SDXL~\cite{podell2023sdxl}, \dalle~\cite{dalle3system,dalle3paper} and Ideogram\footnote{\url{https://ideogram.ai/}}, these models face challenges with prompts requiring numeracy or spatial arrangement capability, for example, Figure~\ref{fig:teaser} (a) showcases a captivating storybook design image generated by \dalle with the text prompt describing the story of the {``\emph{three pigs}''}. We find there are four pigs in the figure, which is not consistent with {``\emph{three pigs}''} in the text prompt.
To overcome these limitations, cutting-edge efforts~\cite{epstein2023diffusion,mou2024diffeditor, shi2023dragdiffusion, mou2023dragondiffusion} have been directed towards developing precise spatial-aware image editing techniques, aiming to bridge the discrepancy between user expectations and initial generation outcomes.

Unlike previous methods~\cite{epstein2023diffusion,mou2024diffeditor, shi2023dragdiffusion, mou2023dragondiffusion} that require combining multiple editing guidance designs for different editing tasks and updating the latent representations through additional backpropagation, we propose a training-free, forward-only, and unified framework for accurate spatial-aware image editing tasks. Our approach transforms most of the representative spatial-aware editing tasks into a two-fold process. This process involves first decomposing the multi-layered latent representations of source images based on precise user instructions and the layer segmentation masks, and then integrating these representations into target images in accordance with an accurate layout arrangement. To ensure accurate spatial-aware editing quality of multiple image layers, we explicitly fuse the multiple layered latents following the target layout arrangement to form the target latent representations. Additionally, we support leveraging the reasoning and visual planning capabilities of GPT-4V~\cite{yang2023dawn} to assist in crafting user instructions and generating (and refining) accurate layout arrangements.

We identify the key challenges in performing the multi-layered latent decomposition and fusion process, and then present three non-trivial technical contributions as follows: 

(i) First, we observe that one of the key challenges in performing the multi-layered latent decomposition lies in generating a high-quality background layer. This layer should not only maintain faithfulness to the original ones but also inpaint the incomplete regions of the decomposed object layers. Instead of applying existing inpainting methods, we introduce a very simple yet more reliable self-attention~\cite{vaswani2023attention} key-masking approach that consistently achieves much better inpainting quality.
(ii) Second, another challenge we need to address is that the inpainted region might suffer from the negative influence of some unrelated areas, leading to artifacts. Therefore, we propose an artifact suppression scheme to further enhance the inpainting quality. (iii) Third, we introduce a unified framework for various image editing tasks by breaking them down into two fundamental sub-tasks: multi-layered latent decomposition and multi-layered latent fusion.

We perform an extensive user study to evaluate the image editing quality of our approach, comparing it to the latest advancements in Self-Guidance~\cite{epstein2023diffusion} and DiffEditor~\cite{mou2024diffeditor}. The outcomes, illustrated in Figure~\ref{fig:user_study}, showcase the win rates across two key dimensions: image quality and editing fidelity. Our findings demonstrate that our method significantly outperforms these two benchmark approaches in various editing tasks, such as object movement and resizing. Additionally, we apply our approach to a range of challenging design image editing tasks, such as object removal, resizing, movement, repetition, flipping, camera panning, zooming out, composing multiple images, and editing typography or decorations, among others. We hope to inspire further developments in more precise spatial-aware image editing technologies.

\begin{figure*}[!htbp]
\centering
\includegraphics[width=0.99\textwidth]{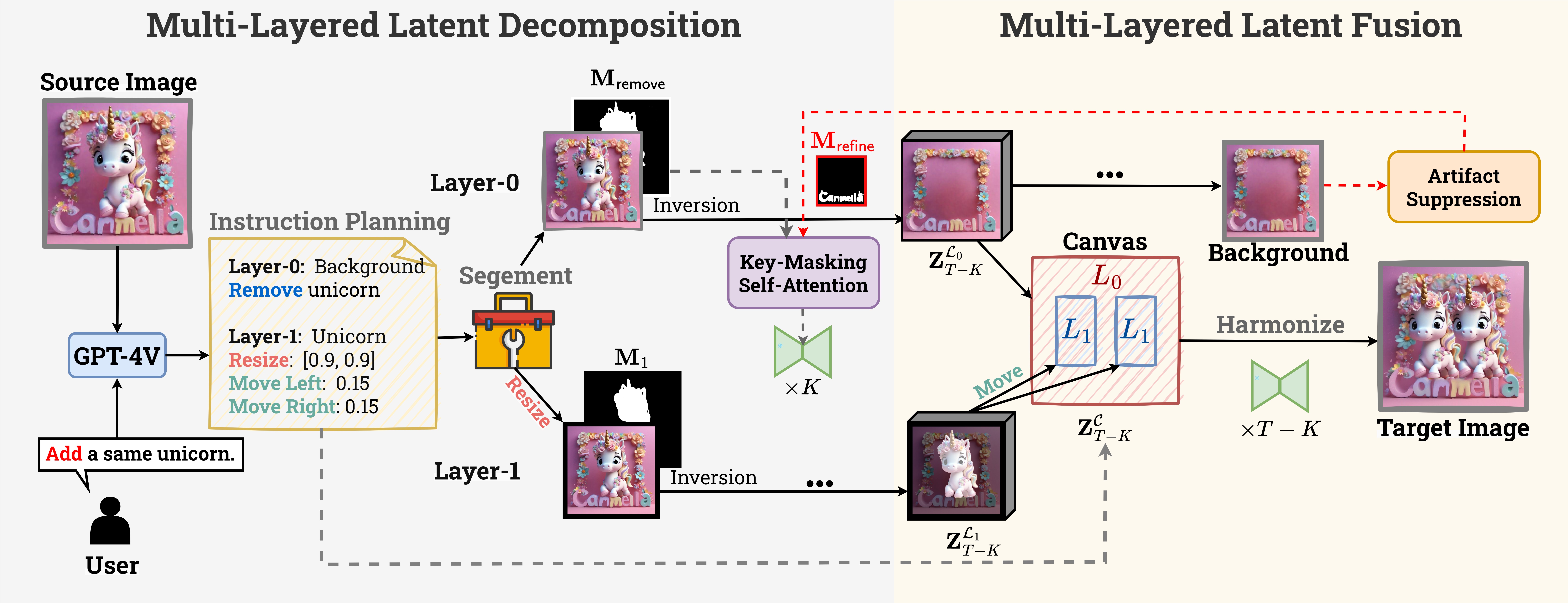}
\vspace{-2mm}
\caption{{
\textbf{Illustrating the overall framework of our approach:} During the multi-layered decomposition stage, given a user's editing instruction and the source image, we first utilize GPT-4V to perform instruction planning, generating a set of detailed layer-wise editing instructions. Then, we segment the source image into multiple image layers, including the background layer that requires additional inpainting, implemented by a novel key-masking self-attention scheme, and the other object layers of the object to manipulate. For the multi-layered fusion stage, We follow the layers' orders and layer-wise instructions sequentially to paste them onto the canvas in latent space. We further apply multiple denoising steps to harmonize the fused multi-layered latent representations. Additionally, we perform artifact suppression to improve the background inpainting quality.
}}
\label{fig:design_edit_framework}
\end{figure*}

%% file: sec/2_relate.tex
\section{Related Work}
\label{sec:related_work}

\subsection{Latent Diffusion Model}
Latent Diffusion Models~\cite{rombach2022highresolution} (LDMs) introduce a groundbreaking approach to the field of generative modeling by operating in a compressed latent space, rather than at the image level. This method accelerates the generation process and reduces computational demands. Recently, large-scale conditional diffusion models~\cite{rombach2022highresolution, podell2023sdxl, saharia2022photorealistic} that adopt the architecture of latent diffusion models and are trained on a large amount of data, can generate images that are both rich in detail and visually appealing. Image editing methods like Blended Latent Diffusion~\cite{avrahami2023blended} demonstrate that operating in the latent space can achieve local image adjustments with faster inference and better precision than operating at the image level~\cite{Avrahami_2022}. In our work, we adopt the state-of-the-art large-scale text-to-image LDMs, Stable Diffusion~\cite{rombach2022highresolution, podell2023sdxl} with U-Net structure~\cite{ronneberger2015unet}, to further explore latent operations for spatial-aware image editing.

\subsection{Guidance-Driven Spatial-aware Image Editing}
Spatial editing involves modifying images by considering the spatial context and relationships within the image. This includes removing, moving, resizing, or adding elements, in contrast to in-place editing methods~\cite{hertz2022prompttoprompt, cao2023masactrl, hertz2024style, brooks2023instructpix2pix}.

Inspired by the classifier guidance strategy on diffusion models, Training-free Layout Control~\cite{chen2023training} and Boxdiff~\cite{xie2023boxdiff} constrain the latent space using position information loss to achieve spatial-aware image generation with layout control. Self-Guidance~\cite{epstein2023diffusion} introduces classifier-guidance into diffusion-based image editing to complete tasks like object movement and resizing. DragonDiffusion~\cite{mou2023dragondiffusion}, inspired by DragGAN~\cite{pan2023drag}, incorporate dragging-based image editing tasks into diffusion models, extending to more spatial-aware editing tasks, such as object movement and resizing with image prompts like object masks. DiffEditor~\cite{mou2024diffeditor} improves DragonDiffusion to achieve state-of-the-art results in accurate image editing tasks.

These guidance-driven methods~\cite{chefer2023attendandexcite, yu2023freedom, yu2023freedom02, epstein2023diffusion, mou2023dragondiffusion, mou2024diffeditor} rely on loss backward, leading to the entanglement among various elements, making it impractical to perform different operations on different objects simultaneously. Ours solves the problem by multi-layer decomposition, utilizing the flexibility of layers to achieve more complex and general editing tasks. On the other hand, loss is a soft constraint that ignores or modifies relative pixel-level features, potentially leading to changes in object and background identity. Our multi-layer fusion strategy directly follows layer-wise editing instructions in latent space. Additionally, our approach facilitates object removal with performance matching that of specifically trained or tuned inpainting models, a capability lacking in guidance-driven methods.

%% file: sec/3_approach.tex
\section{Approach}
\label{sec:approach}

\begin{figure*}[!h]
    \centering
    \includegraphics[width=0.9\linewidth]{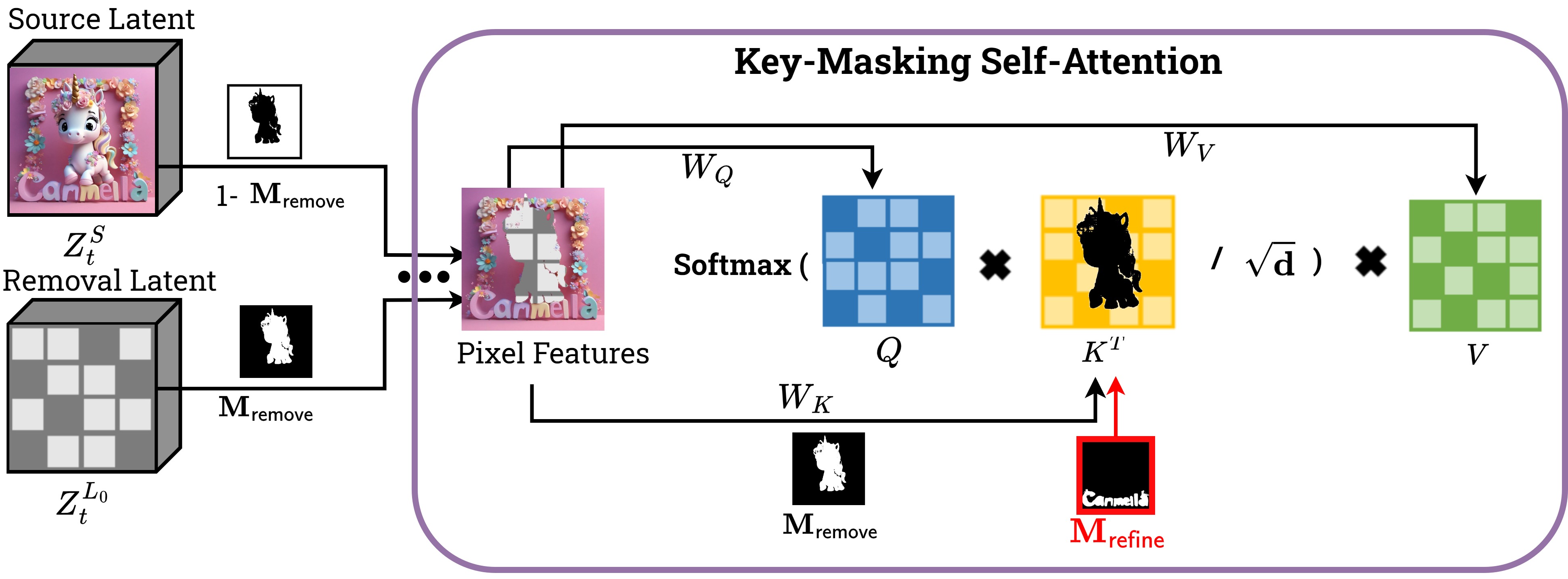}
    \caption{\textbf{Key-Masking Self-Attention Mechanism at time step $\textbf{t}$.} The figure shows the diagram for the removal latent ${\bf Z}_t^\mathcal{S}$ at timestep \(t\). The surroundings of pixel features are kept by the source latent ${\bf Z}_t^\mathcal{S}$. ${\bf M}_\mathsf{remove}$ and ${\bf M}_\mathsf{refine}$ are utilized on key features to reduce attention within the mask.}
    \label{fig:qkv_2}
\end{figure*}

The key idea of our work is at presenting a training-free multi-layered decomposition and fusion framework that can unify various spatial-ware image editing tasks.
First, we explain the detailed design of multi-layered latent decomposition stage that prepares the precise layered-latent representations associated with different objects based on a set of object segmentation masks.
Additionally, we leverage the reasoning and planning capabilities of GPT-4V to automatically transform user editing requests into structured, layer-wise editing instructions by providing several in-context examples.
Second, we demonstrate a multi-layered latent fusion scheme that integrates multiple latent representations in accordance with a target layout canvas, which can be supplied by either human input or GPT-4V. Last, we enhance the harmony of the fused target latent representations by applying additional diffusion steps.
Moreover, we introduce an artifact suppression refinement strategy to check and enhance the effectiveness of the background removal. Figure~\ref{fig:design_edit_framework} illustrates the overall pipeline of our approach.

\subsection{Multi-Layered Latent Decomposition}
\label{sec:fusion_31}
Inspired by the concept of layers in the design domain, we introduce a multi-layered latent decomposition scheme to simplify the complex image editing process into a set of independent, easily manageable layer-wise editing operations for each image layer.
In this study, we conceptualize a ``layer'' as either a singular basic visual element or a collection of multiple visual elements within the source image. Each layer can be independently adjusted, removed, or merged with others, facilitating precise manipulation of the final image composition.

Given a source image and an editing instruction, we need to perform layer-wise editing instruction planning and prepare the multi-layered latent representations. More details are illustrated as follows.

\vspace{1mm}
\noindent\textbf{Layer-wise Editing Instruction Planning} 
The key idea of this step is to leverage the reasoning and planning capabilities of GPT-4V to transform vague user editing instructions into detailed and clear, layer-wise editing instructions. We support two types of spatial editing instructions: ``Resize'', which adjusts size using height and width ratios, and ``Move'', which adjusts position using direction and scale. The layer order depends on the sequence of pasting on the canvas. Layer-0 serves as the background layer, while Layer-1 to Layer-N serve as instance layers.

\vspace{2mm}
\noindent\textbf{Layer-wise Mask Segmentation and Adjustment}
After generating the layer-wise editing instructions, we proceed with layer-wise mask segmentation for two purposes: object removal through the key-masking self-attention scheme and as foundational elements for constructing the layout canvas. An interesting observation we've made is that merely resizing the layer-wise latents can lead to blurring and artifacts. To address this, we resize both the initial image and mask, then encode the resized one into latent space while maintaining the object's central positioning unchanged.

\vspace{2mm}
\noindent\textbf{Key-Masking Self-Attention}
Then we encode the prepared layer image into latent space by inversion technique~\cite{han2023improving}. We introduce a novel key-Masking self-attention scheme within the U-Net structure of the Latent Diffusion Model to remove the regions inside the mask of Layer-0 and maintain the overall harmony of the background. 

Key-Masking Self-Attention applies the removal mask \({\bf{M}}_{\mathsf{remove}}\) to the \textsc{key} features of the self-attention during the initial \(K\) diffusion steps. The computational process is described as follows:
\begin{align}
\label{eq:self_attention_01}
\operatorname{Softmax}\left(\frac{\mathbf{Q}\;((1-\bf{M}_{\mathsf{remove}})\odot\mathbf{K})^{\text{T}}}{\sqrt{d}}\right)\mathbf{V},
\end{align}
where Q, K, V come from the removal latent features \({\bf{Z}}_{T}^{\mathcal{L}_0}\), projected by $W_Q$, $W_K$, $W_V$. 

To preserve the areas outside the mask, we replicate the surrounding features from the source latent \({\bf{Z}}_{t}^\mathcal{S}\) provided by the inversion path. \({\bf{Z}}_{T}^{\mathcal{L}_0}\)  is initialized by \({\bf{Z}}_{T}^{\mathcal{S}}\). As shown in Figure~\ref{fig:qkv_2}, at each denoising timestep \( t \), we update the removal latent \({\bf{Z}}_{t}^{\mathcal{L}_0}\) to retain the latest surrounding features:

\begin{align}
\label{eq:self_attention_02}
   {\bf{Z}}_{t}^{\mathcal{L}_0} = {\bf{Z}}_{t}^{\mathcal{L}_0} \odot {\bf{M}_{\mathsf{remove}}} + {\bf{Z}}_{t}^\mathcal{S} \odot (1-{\bf{M}_{\mathsf{remove}}}).
\end{align}

\begin{figure}[!h]
    \centering
    \includegraphics[width=0.475\textwidth]{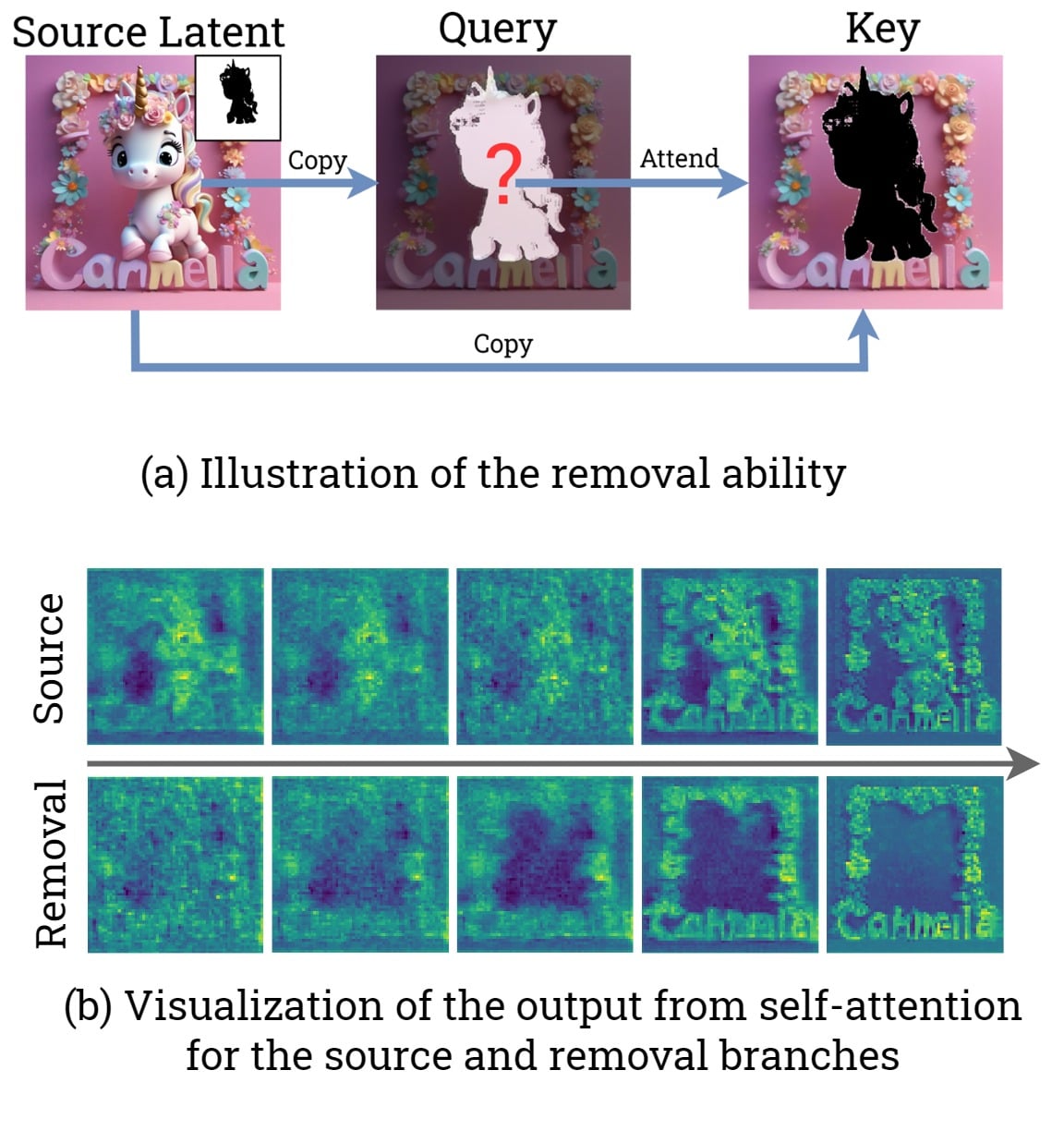}
    \caption{\textbf{Illustrating the Key-Masking Self-Attention Mechanism.} (a) shows that regions inside the mask query only from the regions outside the mask, which are copied from the source latent to complete the information. (b) presents the output heatmaps changing over time from the source and removal latent. The maps come from the first self-attention block at a resolution of $ 64 \times 64$ .}
    \label{fig:vis_self}
\end{figure}

Since attention weights are calculated by matching query and key, if a key is masked, the match degree of any query with this key will be very low. This results in the key's corresponding regions $\mathbf{M}_{\text{remove}}$ not being considered in the weighted sum computation. 
Figure~\ref{fig:vis_self} (a) illustrates that by applying the mask to the key features, we enable the query to \textit{ignore} the regions inside the mask, focusing only on the remaining areas. The regions inside the mask are reconstructed by consulting the remaining areas which are preserved step by step by ${\mathbf{Z}}_{t}^\mathcal{S}$. Figure~\ref{fig:vis_self} (b) visualizes heatmaps of the output features of self-attention from the source latent ${\mathbf{Z}}_{t}^\mathcal{S}$ and the removal latent ${\mathbf{Z}}_{t}^{\mathcal{L}_0}$. We observe that information corresponding to the masked region is suppressed in the final output, receiving a lower attention score compared to the source latent, while ensuring a gradual transition with the surrounding background.

\subsection{Multi-Layered Latent Fusion}
\label{sec:fusion_32}

\begin{table*}[!h]
\begin{minipage}[t]{1\linewidth}{
\tablestyle{10pt}{1.1}
\resizebox{\linewidth}{!}{
\begin{tabular}{l|c|c|c|c}
\textbf{Editing Task} & \textbf{Adjust Image} & \textbf{Remove Mask ${\bf{M}}_{\mathsf{remove}}$} & \textbf{Source, Removal, Target} 
& \textbf{Fusion $t$} \\
\shline
Object Removal & None & $\sum{{\bf{M}}_{\mathsf{obj}}}$
&${\bf{Z}}_{t}^{\mathcal{S}}$, ${\bf{Z}}_{t}^{\mathcal{L}_0}$, ${\bf{Z}}_{t}^{\mathcal{L}_0}$
& None \\
Object Movement & None & $\sum{{\bf{M}}_{\mathsf{obj}}}$
&${\bf{Z}}_{t}^{\mathcal{S}}$, ${\bf{Z}}_{t}^{\mathcal{L}_0}$, ${\bf{Z}}_{t}^{\mathcal{C}}$
& $T-K$ \\
Object Resizing, Flipping & Resize, Flip  & $\sum{{\bf{M}}_{\mathsf{obj}}}$ &${\bf{Z}}_{t}^{\mathcal{S}}$, ${\bf{Z}}_{t}^{\mathcal{L}_0}$, ${\bf{Z}}_{t}^{\mathcal{C}}$ & $T-K$ \\
Camera Panning & Pan and Paste & ${\bf{M}}_{\mathsf{pan}}$ & ${\bf{Z}}_{t}^{\mathcal{S}}$, ${\bf{Z}}_{t}^{{\mathcal{L}_0}}$,
${\bf{Z}}_{t}^{\mathcal{L}_0}$ & None \\
Zooming Out & Zoom and Paste & ${\bf{M}}_{\mathsf{zoom}}$ & ${\bf{Z}}_{t}^{\mathcal{S}}$, ${\bf{Z}}_{t}^{{\mathcal{L}_0}}$,
${\bf{Z}}_{t}^{\mathcal{L}_0}$ & None \\
Occlusion-Aware Editing & None & $\sum_{v_j \in V_i} \operatorname{Move}({\bf{M}}_\mathsf{occlude}; \mathbf{v_j})) $ & ${\bf{Z}}_{t}^{\mathcal{C}}$, $\hat{\mathbf{Z}}_{t}^{\mathcal{C}}$,
$\hat{\mathbf{Z}}_{t}^{\mathcal{C}}$
& $T-K \sim 0 $ \\
Cross-Image Composition & Layout-guided & ${\bf{M}}_{\mathsf{BG}}$ & ${\bf{Z}}_{t}^{\mathcal{BG}}$,
${\bf{Z}}_{t}^{\mathcal{L}_0}$, ${\bf{Z}}_{t}^{\mathcal{C}}$ & $T-K$ \\
\end{tabular}}
\vspace{2mm}
\caption{ \footnotesize \textbf{Unified Overview of Spatial-Aware Image Editing Tasks.} ``Source" represents the initial latent before removal, as defined in Equation~\eqref {eq:self_attention_01} and \eqref{eq:self_attention_02}. ``Removal" refers to the latent to apply key-masking self-attention in Equations~\eqref{eq:self_attention_01} and \eqref{eq:self_attention_02}. "Target" latent is used to decode the final output. ``Fusion step \(t\)" is the range where Equation~\eqref{eq:fusion_01} is implemented.}
\label{tab:editing_unify}}
\end{minipage}
\end{table*}

\noindent\textbf{Instruction-Guided Latent Fusion}
After the first K steps of removal on the background layer \(L_0\), we sequentially paste the prepared layered latent features onto the layout canvas latent \({\bf{Z}}_{t}^{\mathcal{C}}\) at timestep \(T-K\) with layer-wise ``Move" instructions $V_i$. 

Given a two-dimensional operating vector \(\mathbf{v} = (dx, dy)\), we define the operation $ 
\operatorname{Move}(I; \mathbf{v}): B \times C \times H \times W \to B \times C \times H \times W$ as follows:
\[I'(i, j) = \operatorname{Move} ( I;\mathbf{v})(i, j) = I(i - dx, j - dy),
\]
where \(B\) is the batch size, \(C\) is the channel numbers, and \(H\) and \(W\) are the image height and width, respectively. The operation moves the latent features and mask in a specific direction and scales them to achieve object movement.

At timestep $t=T-K$, first initialize layout canvas latent ${\bf{Z}}_{t}^{\mathcal{C}}$ with ${\bf{Z}}_{t}^{\mathcal{L}_0}$, and then for each Layer $L_i, i=1,2,...,N$ and for each operating vector $\mathbf{v_j} \in V_i$, we denote $\hat{\bf{M}}_{i} = \operatorname{Move}({\bf{M}_{i}};\mathbf{v_j})$, and the latent fusion process is described by the following equation:
\begin{align}
\label{eq:fusion_01}
   {\bf{Z}}_{t}^{\mathcal{C}} = 
   {\bf{Z}}_{t}^{\mathcal{C}} \odot (1- \hat{\bf{M}}_{i}) + \operatorname{Move}({\bf{Z}}_{t}^{\mathcal{L}_i};\mathbf{v_j})
   \odot \hat{\bf{M}}_{i}.
\end{align}

\vspace{2mm}
\noindent\textbf{Fused Latent Harmonization}
To enhance edge integration between layers and address abrupt changes at interfaces, we conduct a harmonization process after sequential layering, at the final \(T-K\) denoising steps of the diffusion process. This method refines blending and reduces visual discrepancies at layer boundaries, improving image quality and realism.

\begin{figure}[!h]
\centering
\includegraphics[width=0.47\textwidth]{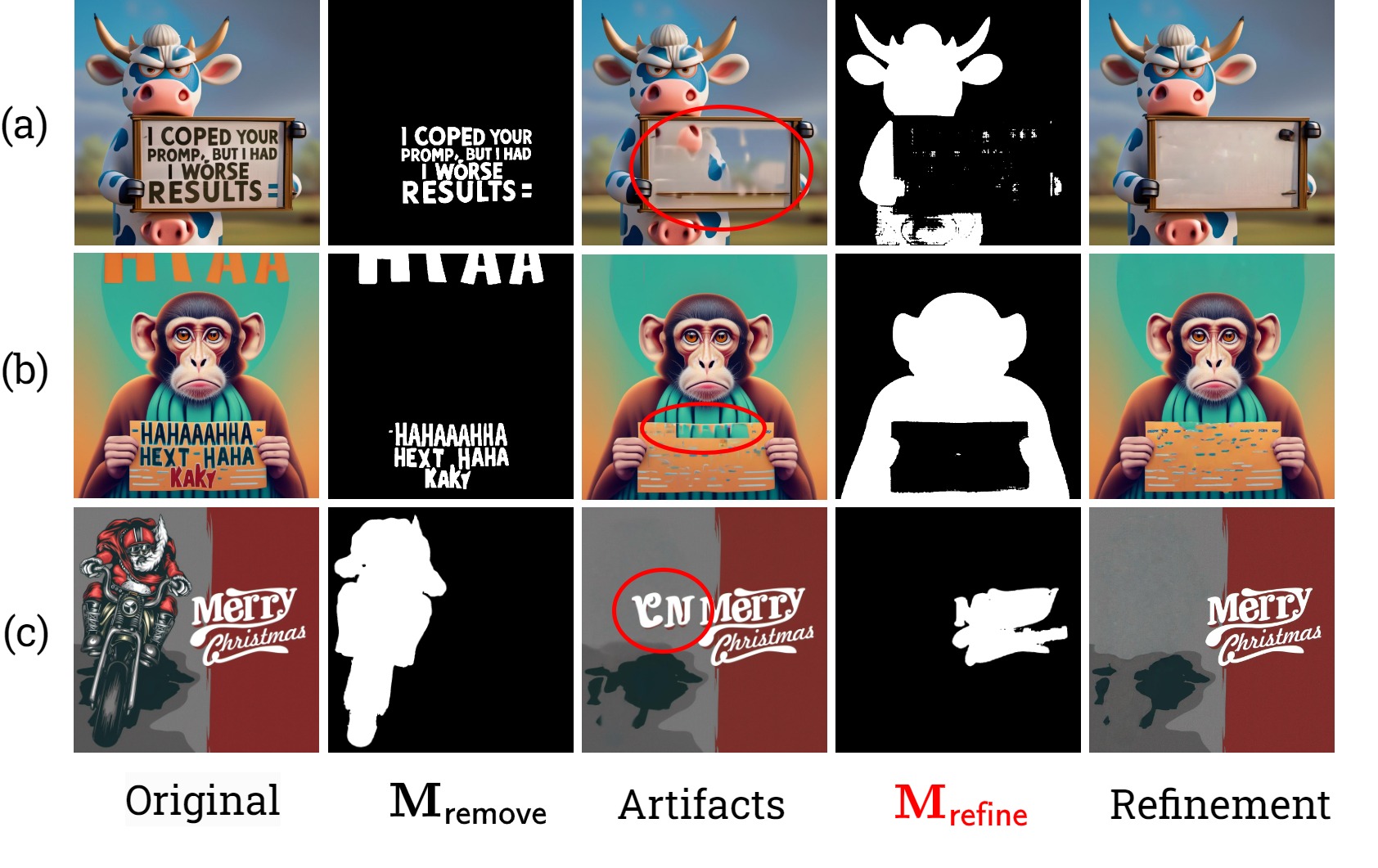}
\caption{
\textbf{Qualitative illustrations of the usage of \(\mathbf{M}_{\mathsf{refine}}\) in artifact suppression refinement.} (a) and (b) show text removal within board elements, while (c) shows the removal of regions near styled text. 
}
\label{fig:artifact}
\end{figure}
\noindent\textbf{Artifact Suppression Refinement} By decoding the inpainted background image, we can check the removal results. 
For examples, as shown in Figure~\ref{fig:artifact} (a) and (b), board elements are common in design images, and removing too much content from the board may result in missing parts.

It's also difficult for the diffusion model to recognize styled typography in some cases, so it tends to extend them in the removal area, as shown in Figure~\ref{fig:artifact} (c). 

To address this issue, we introduce a refinement process, Artifact Suppression. The central idea is to guide the model to avoid focusing on the parts that cause artifacts, which are identified by \({\bf{M}_{\mathsf{refine}}}\). \({\bf{M}_{\mathsf{refine}}}\) is applied together with \({\bf{M}_{\mathsf{remove}}}\) in the Key-Masking Self-Attention Mechanism; it does not affect the latent operations in Equation~\eqref{eq:self_attention_01}. This refinement process enables us to achieve a high success rate in removing content from the source image. The modified key-masking self-attention mechanism is:
\small
\begin{align}
\label{eq.masked_self_attention_artifact}
\operatorname{Softmax}\left(\frac{\mathbf{Q}\;((1-{\bf{M}}_{\mathsf{remove}} - {\textcolor{black}{\bf{M}_\mathsf{refine}}})\odot\mathbf{K})^{\text{T}}}{\sqrt{d}}\right)\mathbf{V}.
\end{align}

\subsection{Unifying Spatial-aware Image Editing Tasks}

Sections~\ref{sec:fusion_31} and \ref{sec:fusion_32} present a general framework for multi-layered representation in image editing. With this framework, we can unify various basic spatial-aware editing operations along with their extensions in Table~\ref{tab:editing_unify}: The removal of masked regions from the ``Source" latent is achieved by applying Key-Masking Self-Attention to the "Removal" latent, as described in Equation~\eqref{eq:self_attention_01} and \eqref{eq:self_attention_02}, thereby enabling multi-layered decomposition. Multi-layered latent fusion is executed by applying Equation \eqref{eq:fusion_01} to the ``Canvas" latent ${\bf{Z}}_{t}^{\mathcal{C}}$.

\vspace{2mm}
\noindent\textbf{Object Removal, Movement, Resizing and Flipping} These are basic editing operations. Resizing and flipping require an additional layer for image-level adjustments to the source image before encoding. Movement is executed during the fusion stage.  The ${\bf M}_\mathsf{remove}$ is the union of masks for all objects needing manipulation, denoted as $\sum{{\bf{M}}_{\mathsf{obj}}}$.

\begin{figure}[!h]
    \centering
    \includegraphics[width=0.475\textwidth]{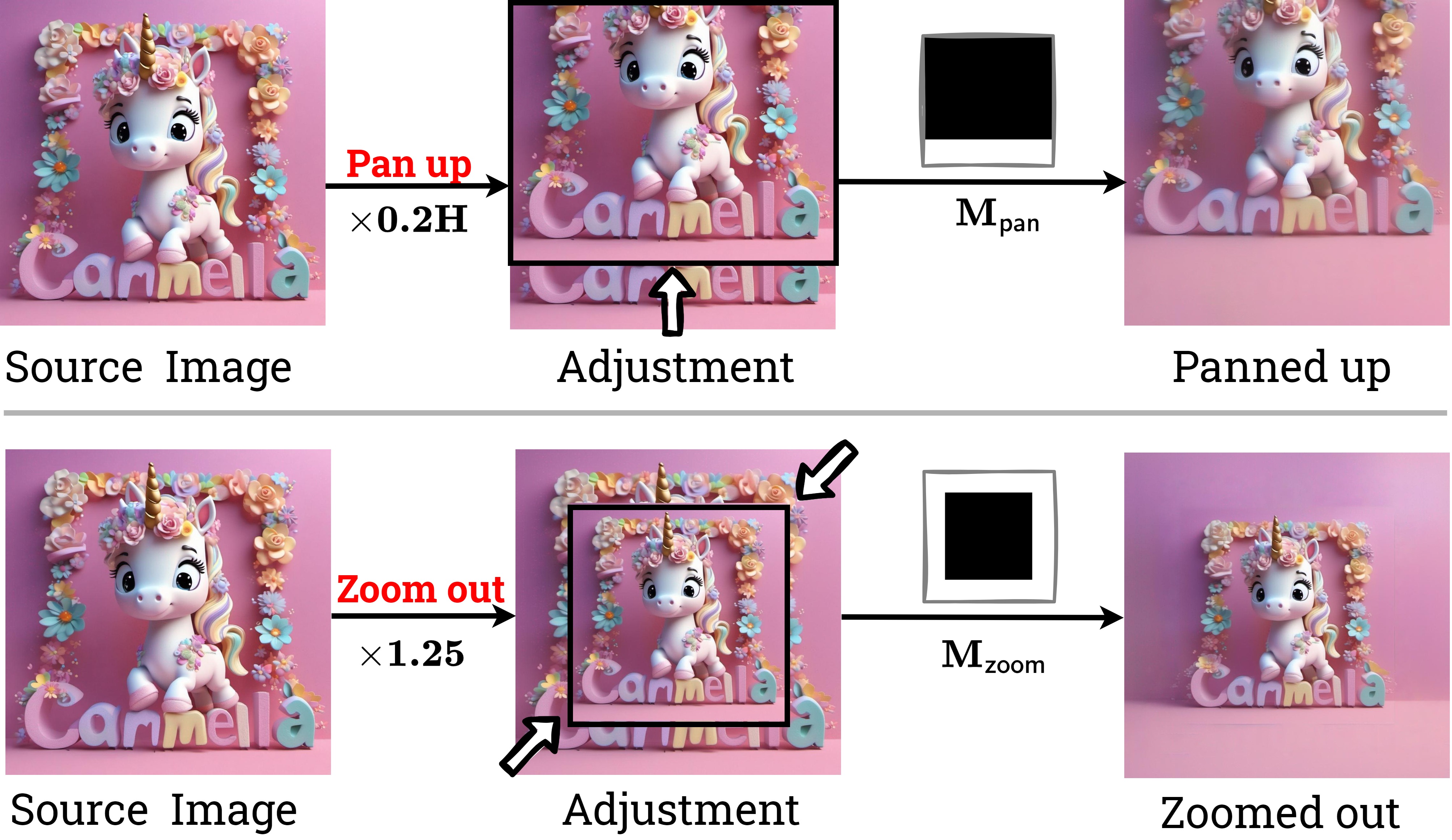}
    \caption{\textbf{Illustration of the mask usage in camera panning and zooming out tasks.} The figure presents two cases of image adjustment and the formation of their related masks.}
    \label{fig:new_pan}
\end{figure}

\vspace{2mm}
\noindent\textbf{Camera Panning and Zooming Out} By adjust the initial image and generating two specific masks, we can convert the tasks of camera panning and zooming out into a removal task. We pan or zoom the source image and paste it onto the original canvas to initialize the removal regions with its adjacent areas, ensuring smooth transitions and color consistency. As shown in Figure~\ref{fig:new_pan}, regions corresponding to the original image are set to 0, and the remaining regions needing completion are set to 1. At the $T \sim T-K$ Removal Stage in Equations~\eqref{eq:self_attention_01} and \eqref{eq:self_attention_02}, we simply replace the $\bf{M}_{\mathsf{remove}}$ with $\bf{M}_{\mathsf{pan/zoom}}$. 

\vspace{2mm}
\begin{figure*}[!h]
\begin{minipage}[h]{1\linewidth}
\begin{subfigure}[b]{1\textwidth}
\centering
\includegraphics[width=0.97\textwidth]{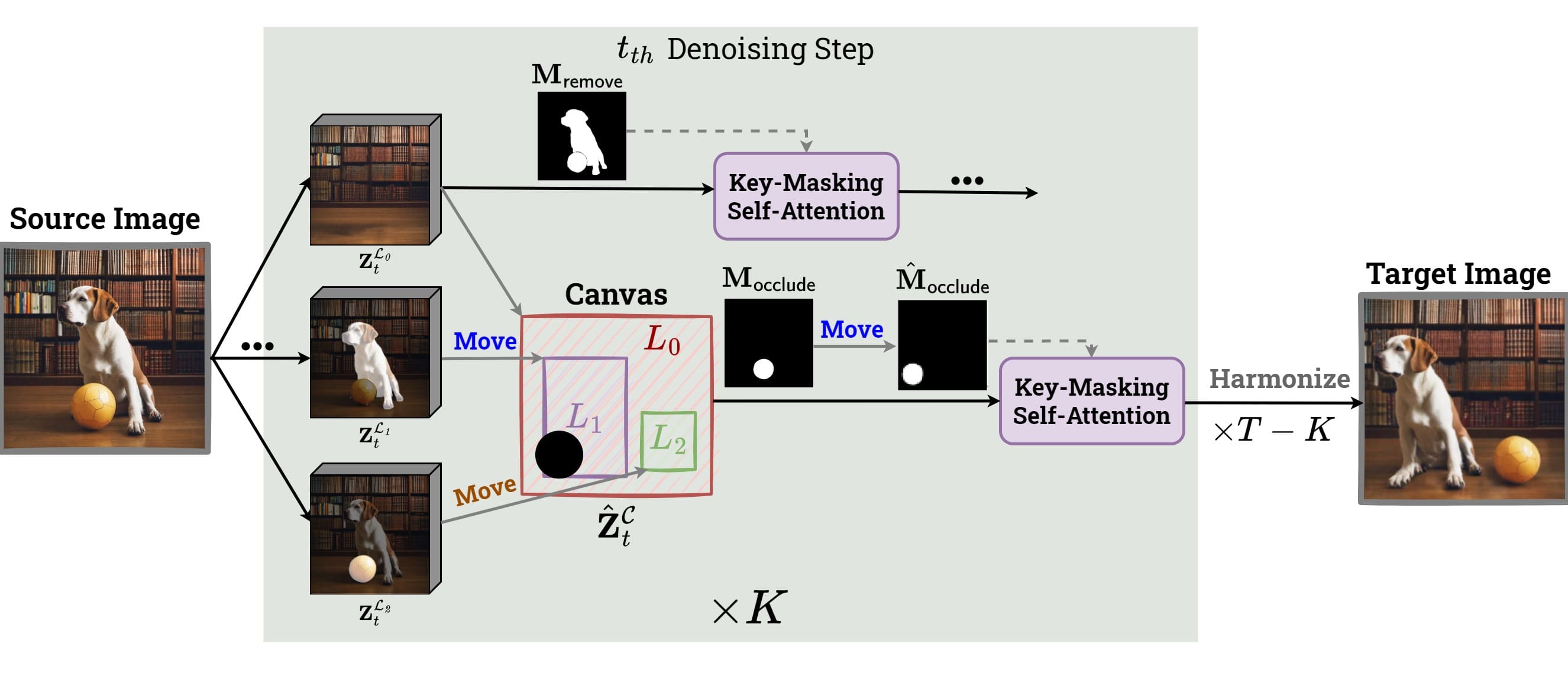}
\end{subfigure}
\end{minipage}
\caption{
\textbf{Illustrating the Integrated Decomposition-Fusion Technique in occlusion-aware object editing at timestep \(t\).} To relocate the dog and ball and inpaint the occluded dog leg, we conduct Key-Masking Self-Attention twice on the background latent ${\bf{Z}}_{t}^{\mathcal{L}_0}$ and the canvas latent \({\bf Z}_t^{\mathcal{L}_0}\) respectively. \(\hat{\bf M}_{\mathsf{occlude}}\) represents the moved \({\bf M}_\mathsf{occlude}\) with the occluded Layer-1 ${\bf{Z}}_{t}^{\mathcal{L}_1}$. The target latent is the new canvas removal latent \(\hat{{\bf Z}}_t^\mathcal{C}\).}
\label{fig:occlude}
\end{figure*}

\begin{table*}[!h]
\centering
\resizebox{1\linewidth}{!}{
\begin{tabular}{cccccccccc}
\textbf{Methods} & L1$\downarrow$ & L2$\downarrow$ & CLIP-I$\uparrow$ & DINO$\uparrow$ & CLIP-T$\uparrow$ & FID$\downarrow$ & LPIPS$\downarrow$ & train or finetune & base model\\ 
\shline
Lama & \textbf{0.014} & \textbf{0.004} & \textbf{0.985} & \textbf{0.979} & \textbf{0.307} & \textbf{25.077} & \textbf{0.053} & $\checkmark$ & - \\  
ControlNet-inpainting & \textcolor{red}{0.025} & 0.010 & 0.951 & 0.905 & 0.300 & 59.322 & 0.095 & $\checkmark$ & SD-v1.5\\
SDXL-inpainting & 0.030 & \textcolor{red}{0.005} & \textcolor{blue}{0.968} & \textcolor{blue}{0.959} & \textcolor{blue}{0.303} & \textcolor{blue}{44.621} & \textcolor{red}{0.069} & $\checkmark$ & SDXL-1.0\\
Uni-paint & 0.064 & 0.020 & 0.920 & 0.837 & 0.298 & 103.846 & 0.156 & $\checkmark$ & SD-v1.4\\
Ours & \textcolor{blue}{0.028} & \textcolor{blue}{0.007} & \textcolor{red}{0.969} & \textcolor{red}{0.971} &  \textcolor{red}{0.306} & \textcolor{red}{38.883} & \textcolor{blue}{0.070} & $\times$ & SDXL-1.0 \\ 
\end{tabular}
}
\vspace{-2mm}
{\captionsetup{hypcap=false}  
\captionof{table}{\textbf{\footnotesize{Quantitative study on the MagicBrush test set for the mask-guided object removal task.}} \textbf{Bold}, \textcolor{red}{Red} and \textcolor{blue}{Blue} represent the top-$3$ results. Our method is the only one that does not require training or finetuning, and it achieves results comparable to SDXL-Inpainting across 7 metrics in 51 examples. Other methods are specifically trained or fine-tuned for mask-guided image inpainting.}
\label{magic}
}
\end{table*}

\noindent\textbf{Occlusion-Aware Object Editing} 
Note that objects often do not appear completely in the source image. For example in Figure~\ref{fig:occlude}, one of the dog's legs is occluded by the ball. Direct relocation results in incomplete failure. We present a novel strategy called \textit{Integrated Decomposition-Fusion Technique}, making full use of the inpainting ability of the Key-Masking Self-Attention. 

The illustration pipeline is shown in Figure~\ref{fig:occlude}. For every iteration of the first K diffusion steps, besides the background removal on ${\bf{Z}}_{t}^{\mathcal{L}_0}$, we first perform the fusion operation on the canvas latent with Equation \eqref{eq:fusion_01}, and then we introduce a new mask ${\bf{M}_\mathsf{occlude}}$, which in this example is the initial ball mask. In this case, the removal latent is represented by the canvas latent $\hat{\mathbf{Z}}_{t}^{\mathcal{C}}$, which is guided by the source latent ${\mathbf{Z}}_{t}^{\mathcal{C}}$:
{
\small 
\begin{align}
\label{eq.masked_self_attention}
   \hat{\mathbf{Z}}_{t}^{\mathcal{C}} = \hat{\mathbf{Z}}_{t}^{\mathcal{C}} \odot \hat{\bf{M}}_\mathsf{occlude} + {\bf{Z}}_{t}^\mathcal{C} \odot (1-\hat{\bf{M}}_\mathsf{occlude}).
\end{align}
}

We denote $\hat{\bf{M}}_\mathsf{occlude} = \sum_{v_j \in V_i} \operatorname{Move}({\bf{M}}_\mathsf{occlude}; \mathbf{v_j})$ to represent the sum of masks after moving with the occluded Layer-i $L_i$. The Key-Masking Mechanism is to replace ${\bf{M}}_\mathsf{occlude}$ with $\hat{\mathbf{M}}_\mathsf{remove}$ in Equation~\eqref{eq:self_attention_01} on canvas latent $\hat{\mathbf{Z}}_{t}^{\mathcal{C}}$.
Integrated Decomposition-Fusion Technique is a more general fusion strategy and in non-occluded image editing contexts, it equals the one-step fusion at $t=T-K$, and the latter has a lower computational cost.

\vspace{2mm}
\noindent\textbf{Cross-Image Composition} Our approach can support cross-image composition by encoding a background reference image (${\bf{Z}}_{t}^{\mathcal{BG}}$) and a set of foreground images. The layered instructions and order are given by the new layout design.

%% file: sec/4_experiment.tex
\section{Experiment}
\label{sec:experiment}

\noindent \textbf{Implementation Details} We made structural modifications to SDXL-1.0~\cite{podell2023sdxl} using the frozen weights and generated images at a resolution of 1024 $\times$ 1024. As a latent diffusion model, the resolution of SDXL-1.0's latent space is 128 $\times$ 128. We adopt the state-of-the-art diffusion inversion technique, Proximal-Guidance~\cite{han2023improving}, to invert the source image into latent space and utilized a 50-step DDIM~\cite{song2020denoising} denoising procedure, which means $T=50$. We selected the most effective value for $K$, which is $K=40$. The key-masking self-attention is applied across all 70 self-attention blocks in SDXL-1.0, with a range of $[50 \sim 10]$.

\begin{figure*}[!h]
    \centering
    \includegraphics[width=\textwidth]{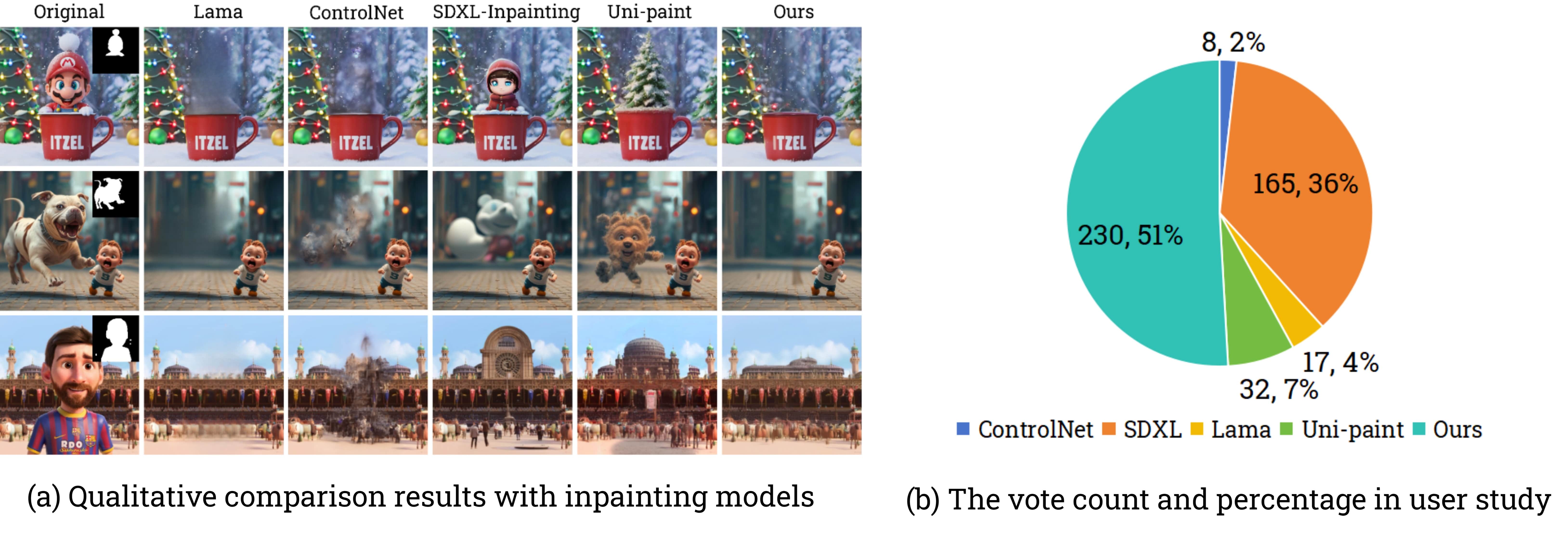}
    \vspace{-2mm}
    \caption{\textbf{Comparison with other mask-guided inpainting models.} (a) shows qualitative the comparison of large object removal ability, with our method not causing obvious blurriness or filling the removed area with unrelated elements. (b) shows the user study results of 452 votes from 113 users, with our method achieving a 51\% preference percentage.}
    \label{fig:removal_cmp}
\end{figure*}

\begin{figure*}[!h]
    \centering
    \includegraphics[width=\textwidth]{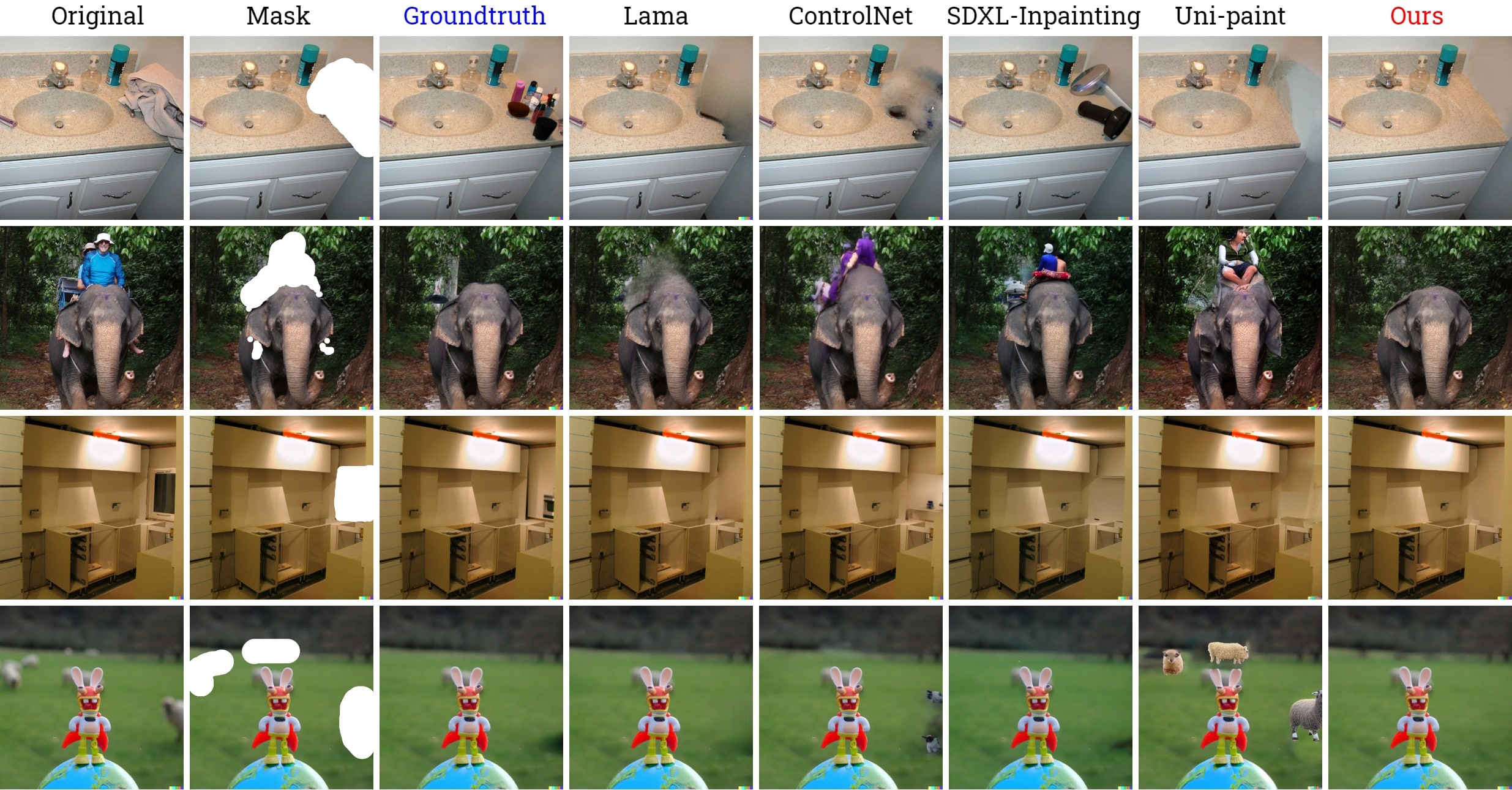}
    \caption{\textbf{Qualitative comparison on the \textsc{MagicBrush} dataset.} We chose the mask-provided instruction-guided removal tasks to evaluate the inpainting ability of our method. The third column shows the removal results provided by DALL$\cdot$ E2 serving as ground truth. We compare the results of LaMa, ControlNet-Inpainting, SDXL-inpainting, Uni-paint with ours.}
    \label{fig:magicbrush}
\end{figure*}

\subsection{Comparison to State-of-the-art}
\label{subsec:cmp}
\vspace{2mm}\noindent\textbf{Object Removal}
We compare the removal ability of our methods with other 5 methods specifically designed for inpainting tasks: Lama~\cite{suvorov2022resolution}, ControlNet-inpainting~\cite{zhang2023adding}, SDXL-inpainting, and Uni-paint~\cite{Yang_2023} on the MagicBrush benchmark~\cite{zhang2023magicbrush}. In our experiments, we utilize data with instructions that are only about removal, with a total of 51 examples, each including a ground truth image generated by DALL$\cdot$ E 2~\cite{ramesh2022hierarchical} for evaluation.

We evaluated the performance of our method and mask-guided image editing method on 7 metrics. L1 and L2 are used to gauge the pixel-level difference between the target image and the ground truth. CLIP-I~\cite{radford2021learning} and DINO~\cite{chen2023subjectdriven} are used to assess image quality, and CLIP-T is used to test text-image alignment. We also utilize the LPIPS~\cite{zhang2018unreasonable} to measures perceptual differences at a patch level, and FID~\cite{heusel2018gans} to assess the similarity between the distributions of real and generated images' features. The quantitative results in Figure~\ref{magic} show the absolute strength of Lama, which is specifically trained for inpainting. Ours achieves results comparable to SDXL-Inpainting across 7 metrics but do not need any finetuning. 

Users were asked to choose the best based on clarity, part restoration, and edge quality. We received 452 votes from 113 users, and the results are shown in Figure~\ref{fig:removal_cmp} (b), which demonstrates the superior performance of our method. Note that although LaMa performs best in benchmark tests, it produces noticeable blurring artifacts when removing large areas, as shown in the second column of Figure~\ref{fig:removal_cmp} (a). This is why it received a low vote count in user studies.

\begin{figure*}[!h]
    \centering
    \includegraphics[width=0.9\linewidth]{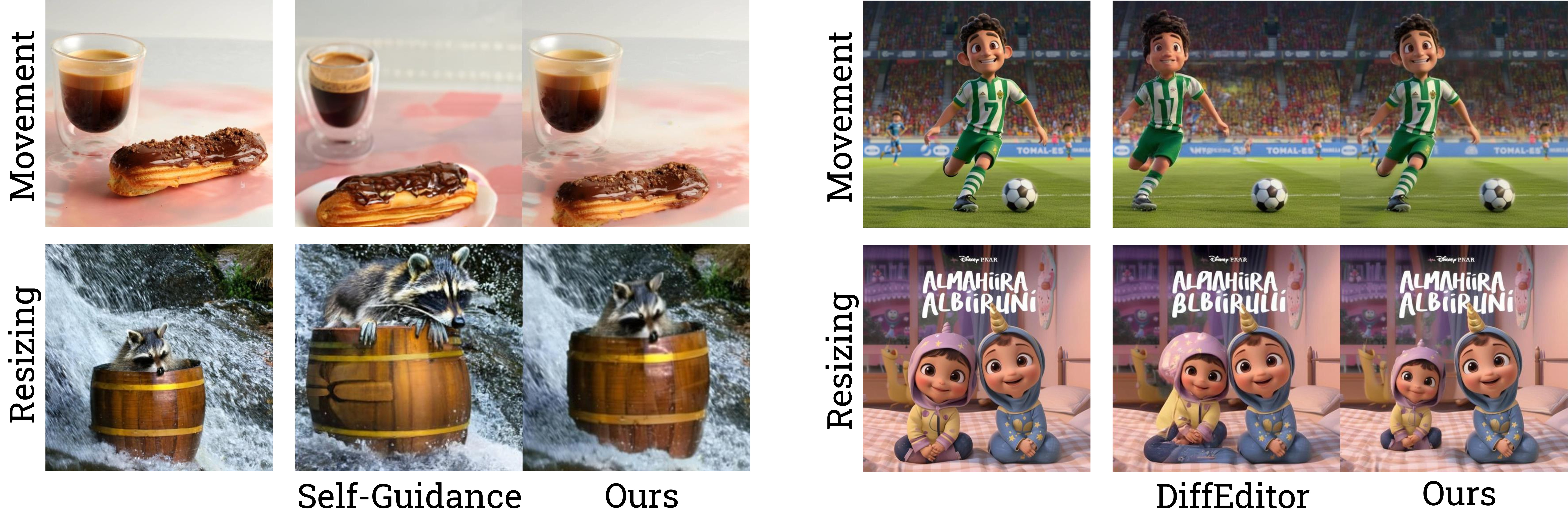}
    \caption{\textbf{More qualitative comparisons with Self-Guidance and DiffEditor.} We conduct single-object editing tasks for movement (the first row) and resizing (the second row). The results in (a) come from the initial paper on Self-Guidance.}
    \label{fig:self_cmp}
\end{figure*}

\begin{figure*}[!h]
    \centering
    \includegraphics[width=\textwidth]{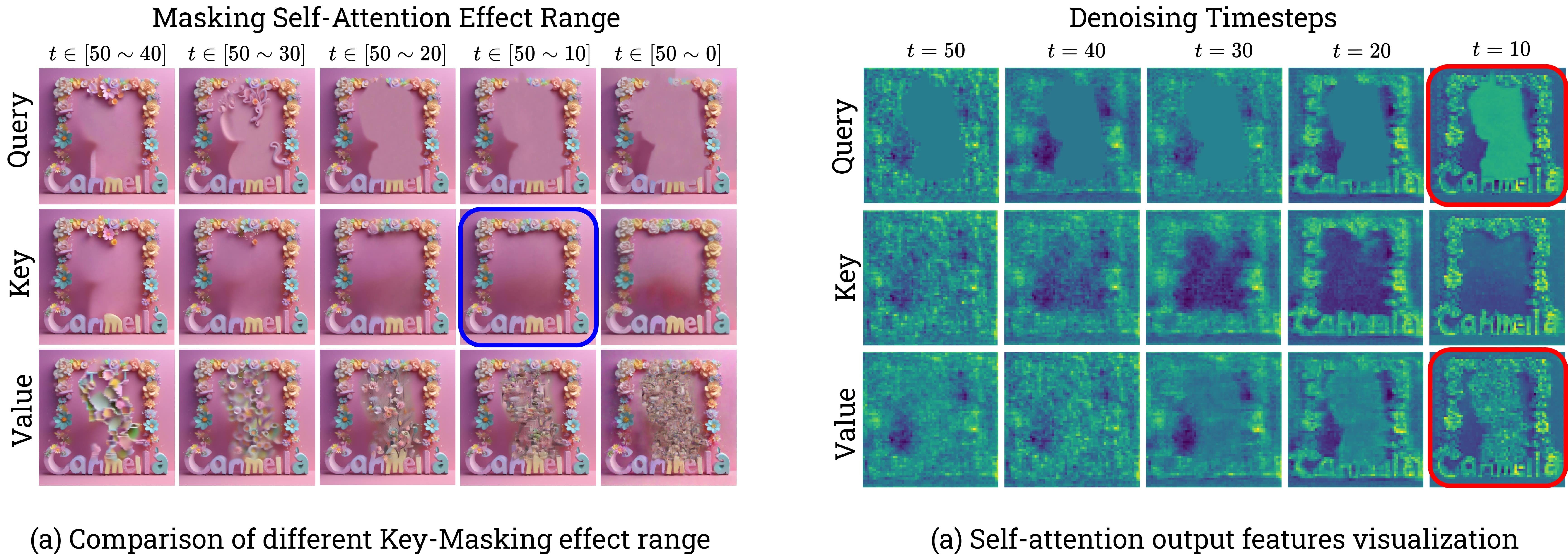}
    \caption{\textbf{The ablation study of different mask placements and effect range of self-attention.} (a) demonstrates the removal results under different masking effect ranges. (b) visualizes the self-attention output at different timesteps under the effect range [$50 \sim 10$], same with the settings highlighted \textcolor{blue}{blue} box in (a).}
    \label{fig:qkv}
\end{figure*}

\begin{figure*}[!h]
    \centering
    \includegraphics[width=\linewidth]{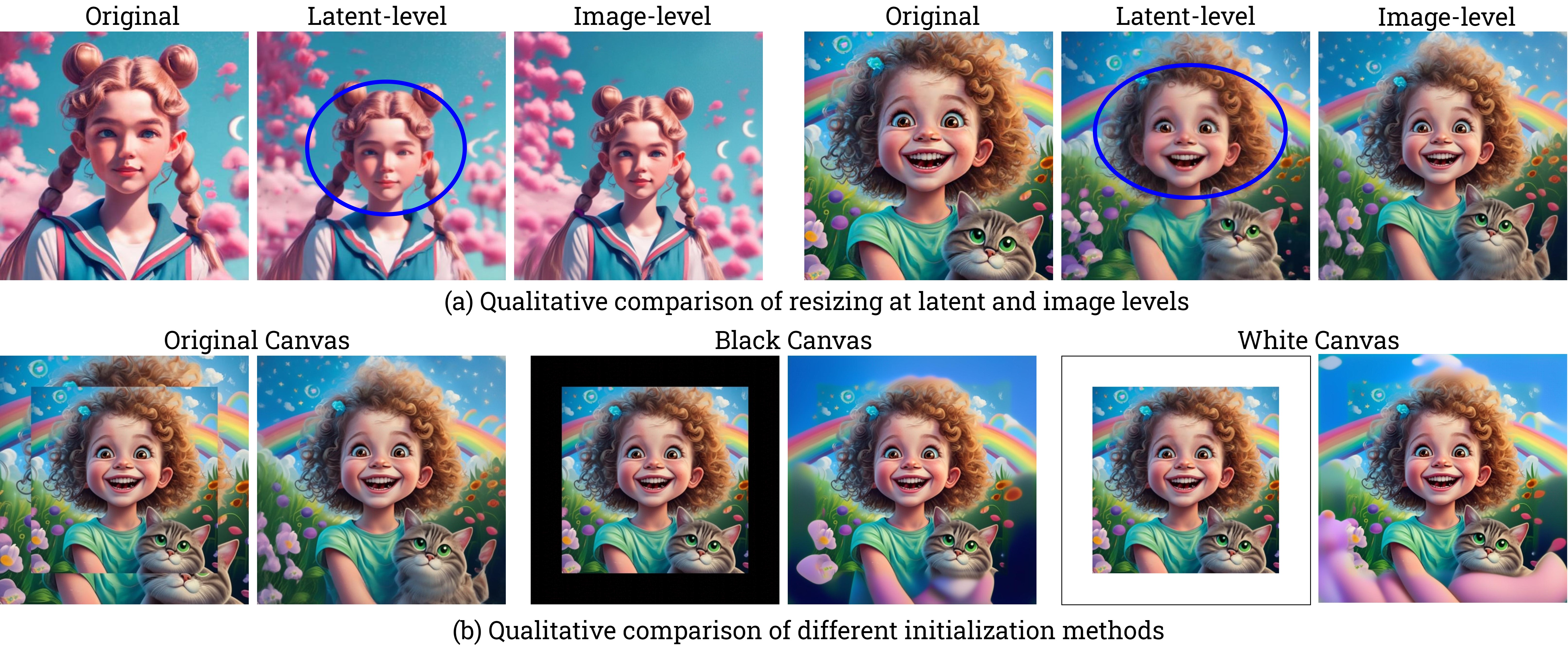}
    \caption{\textbf{The ablation study with zooming out task.} (a) illustrates the different resizing positions at the image level and latent level. (b) shows the different initialization methods with the original image, black canvas, and white canvas.
    }
    \label{fig:resizing_cmp}
\end{figure*}
\vspace{-2mm}

\vspace{2mm}\noindent\textbf{Object Spatial-aware Editing} We present further qualitative comparisons of single-object resizing and movement capabilities between Self-Guidance~\cite{epstein2023diffusion}, DiffEditor~\cite{mou2024diffeditor}, and our method in Figure~\ref{fig:self_cmp}. Our method demonstrates better inpainting performance and superior editing accuracy in preserving the identity of the object and the background, especially with text and large objects. Note that for the other two methods, it is challenging to remove objects or edit two or more objects with different instructions in a single round, such as swapping.

\subsection{Ablation Study} 
\label{subsec:ablation}

\noindent\textbf{Effect Range of Key-Masking Self-Attention} By implementing Equation~\eqref{eq:self_attention_02} across the entire range [$50 \sim 0$] to maintain the surrounding features consistent with the source image, we investigate which effect range results in the most effective removal. As illustrated in the second row of Figure~\ref{fig:qkv} (a), significant removal can be achieved within the first 10 steps of key-masking, while the range [$50 \sim 10$] more effectively integrates the edges and blends better with the background. Therefore, we use $K=40$ as our optimal setting across all editing tasks.

\vspace{2mm}\noindent\textbf{Mask Positioning within Self-Attention Mechanisms} We compare the results of different mask positioning for removal, as shown in Figure~\ref{fig:qkv} (a). Masking the query tends to blur the removal area. As highlighted by the red boxes in Figure~\ref{fig:qkv} (b), the significance of the masked area diminishes during the attention calculation, leading to a reduction in the clarity of the corresponding pixels. Applying masking to the value damages or disrupts the pixels within the masked area, as incorrect values are assigned to these pixels, distorting the generated image. It is important to note that masking on the key results in a smooth transition in the self-attention output around the masked area, ensuring a more coherent integration with the surrounding information.

\vspace{2mm}\noindent\textbf{Layer-wise Size Adjustment: Latent vs. Image} We adjust image size at the multi-layered decomposition stage, which is different from the position adjustment at the latent level. Here, we compare two resizing methods and use the zooming out task as an example, which can be considered as a global resizing of the original image. As shown in Figure~\ref{fig:resizing_cmp} (a), the details of the girls' faces are altered when resizing at the latent level (highlighted in blue circles) and tend to become blurry, losing detail, resulting in inconsistency between the original and target images, thus compromising accuracy. 

There are two main reasons. First, the resolution difference: the resolution at the image level is $1024 \times 1024$, whereas at the latent level, it is $128 \times 128$. The resolution is much lower at the latent level, resulting in significantly lower information density. Second, information loss and compression: resizing in the latent space essentially means further manipulating representations that have already been abstracted and compressed by the model. Due to the nonlinearity and complexity of this process, each feature point represents more abstract, higher-level information about the image. 

Therefore, resizing at the latent level is more likely to result in the loss of these higher-level features, leading to a loss of detailed information. Due to the additional encoding and inversion costs associated with adjustments at the image level, we choose to perform position adjustments at the latent level, which has yielded satisfactory results.

\vspace{2mm}\noindent\textbf{Extra Canvas Initialization: Original Canvas vs. Black Canvas vs. White Canvas} For camera panning and zooming out tasks, we first pan or zoom the initial image to the target position, and then paste it onto the initial image. This approach effectively initializes the regions within the mask using the surrounding areas. Initializing with the original canvas actually provides the model with clues and expected content to fill, allowing for the generation of details consistent with the surrounding environment. Additionally, the model attempts to maintain this coherence, producing content that matches the original image. 

As demonstrated in Figure~\ref{fig:resizing_cmp} (b), we explore two other initialization methods for the zooming out task: black canvas and white canvas. It's observed that the first method inpaints the unknown areas with consistent, intricate details similar to the surroundings, effectively extending clouds, flowers, and the arm of the girl. However, the second and third methods result in inpainting regions that are disjointed and even discordant. This occurs because the model receives a "blank signal," and relying solely on the self-attention mechanism's queries makes it challenging to generate complex details closely connected to the original image.

\subsection{More Qualitative Results}

\begin{figure*}[!h]
    \centering
    \includegraphics[width=0.98\textwidth]{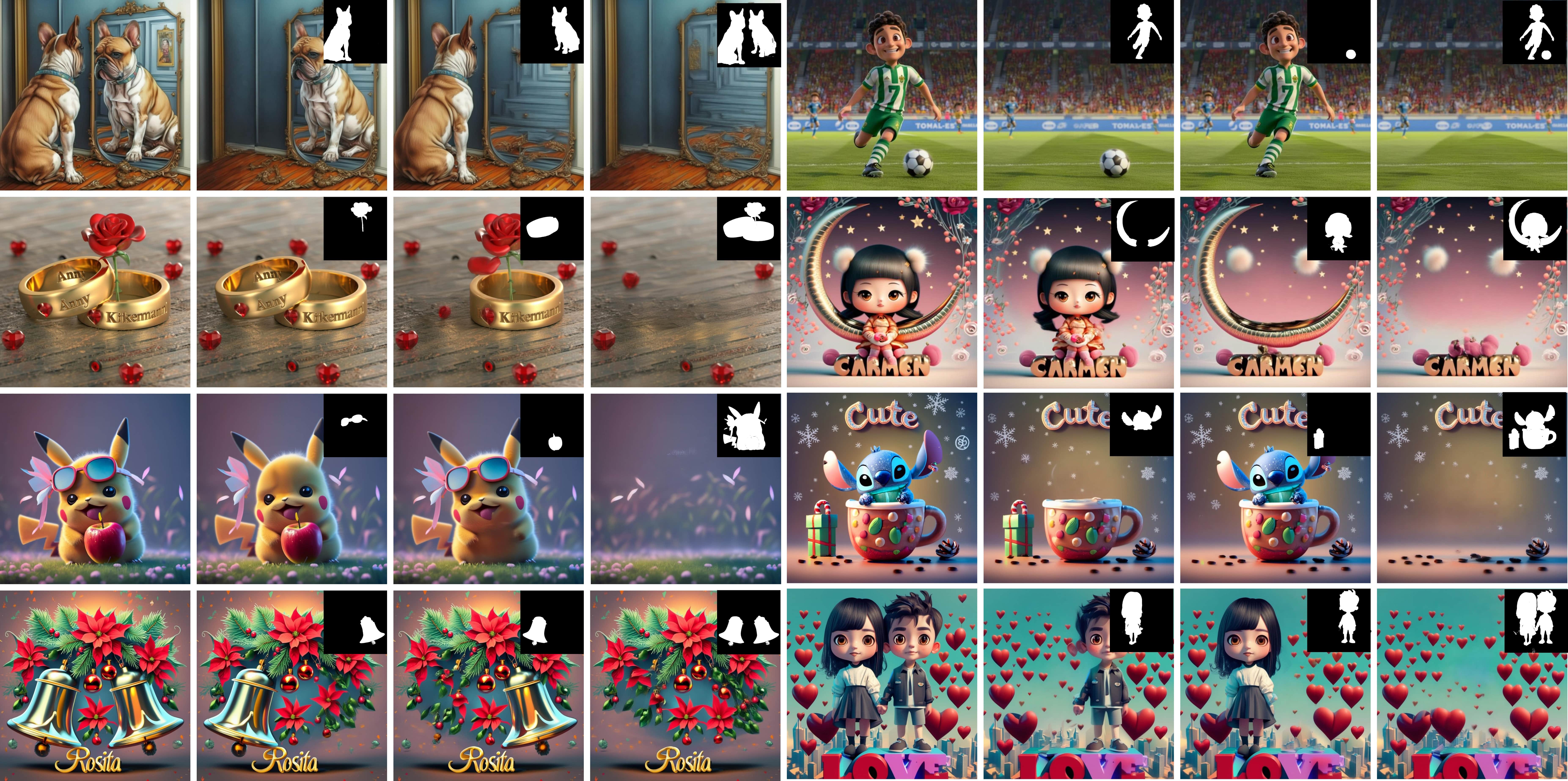}
    \caption{\textbf{Qualitative results of applications on design images.} The figure shows object removal results for single (the second and third columns) and multiple objects (the fourth column), covering the removal of both large and small areas.}
    \label{fig:removal}
\end{figure*}

\begin{figure*}[!h]
    \centering
    \includegraphics[width=0.98\textwidth]{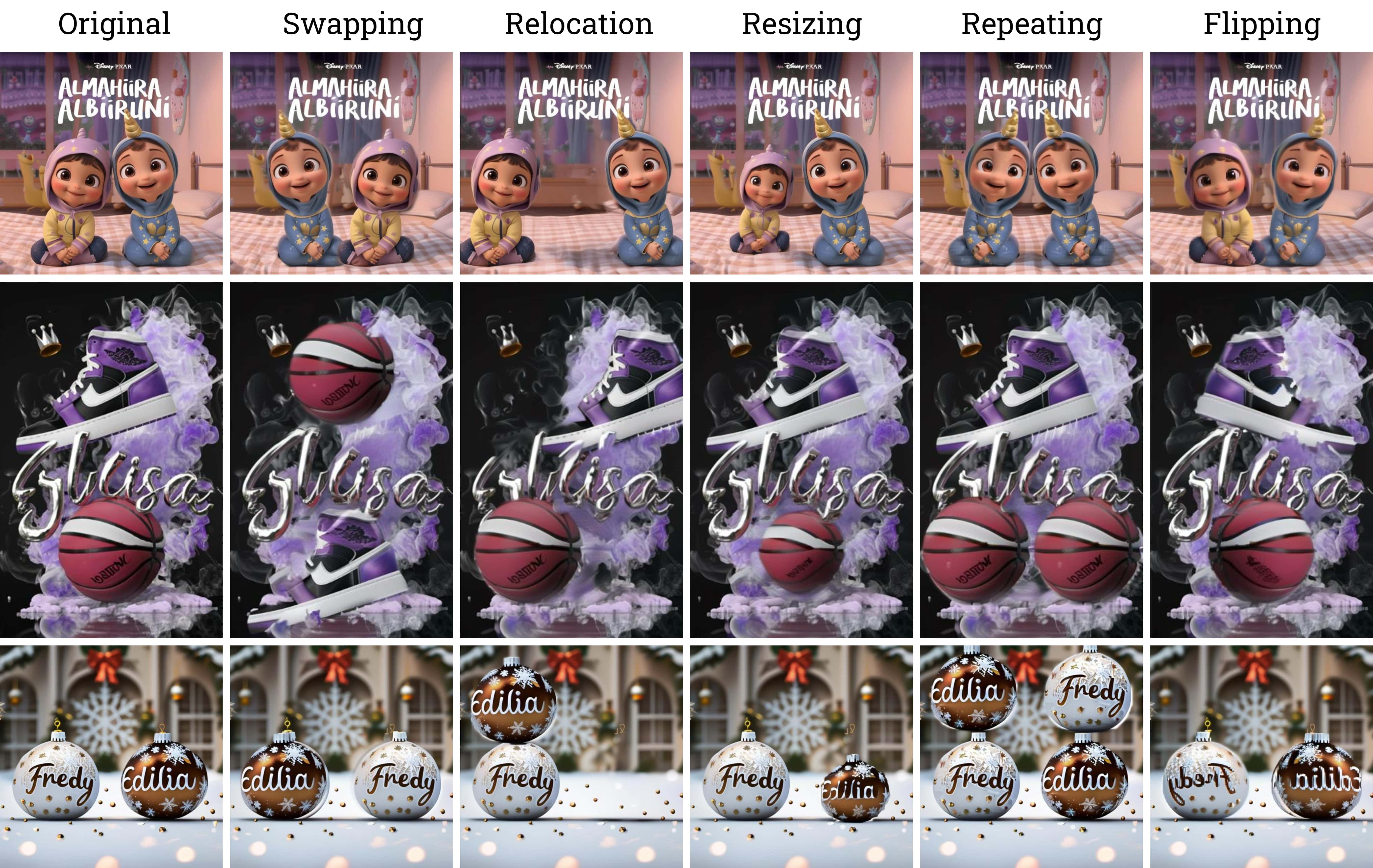}
    \caption{\textbf{Qualitative results of applications on design images.} The figure shows basic editing operations on two-object-centric design images with text elements.}
    \label{fig:relocation}
\end{figure*}

\begin{figure*}[!h]
    \centering
    \includegraphics[width=0.8\textwidth]{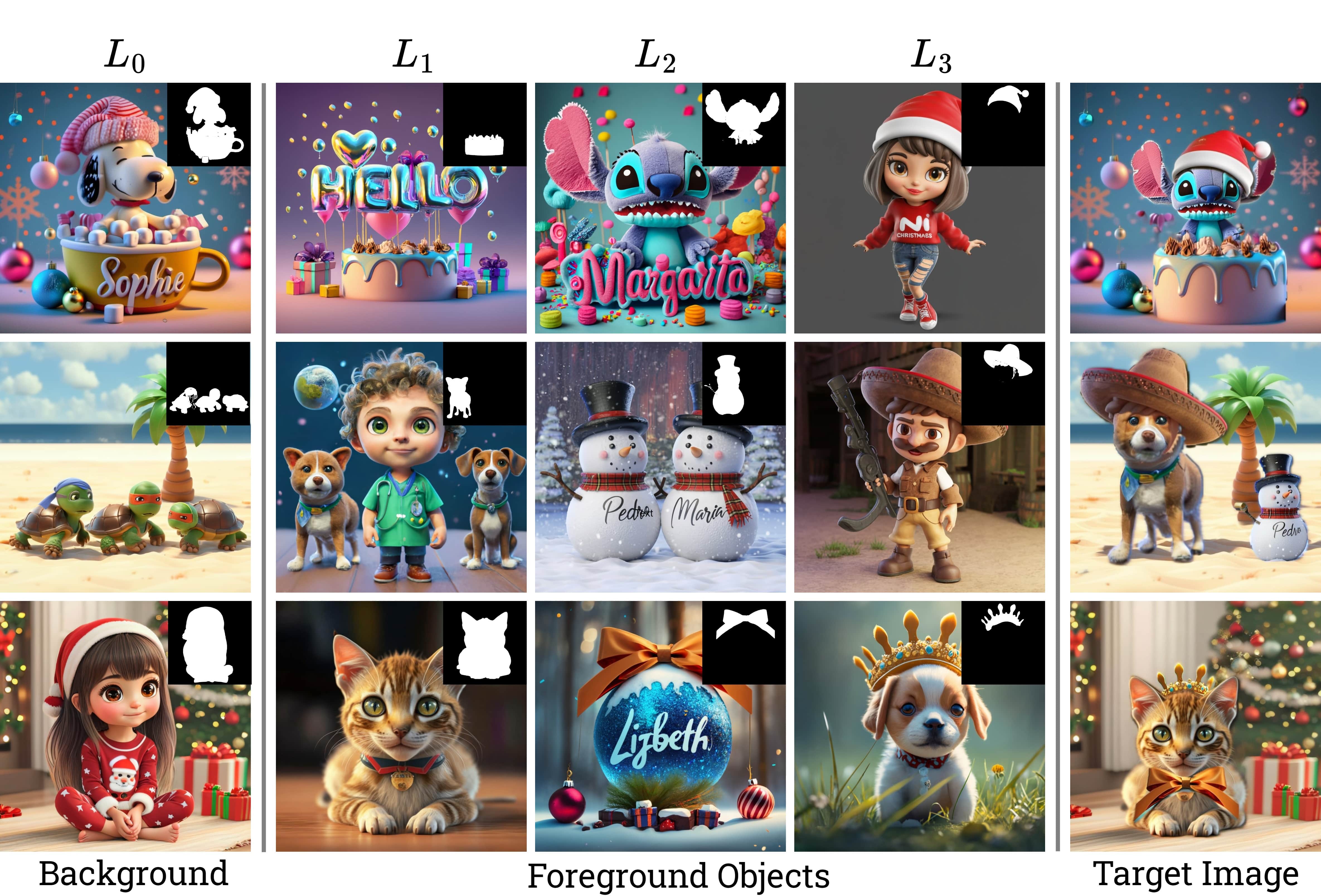}
    \caption{\textbf{Qualitative results of applications on design images.} The figure displays the background and foreground objects, along with their layer orders}
    \label{fig:cross}
\end{figure*}

\noindent\textbf{Multi-Object Complex Editing} We demonstrate our multi-object editing ability with complex operations such as removal in Figure~\ref{fig:removal}, swapping, relocation, resizing, addition, and flipping in Figure~\ref{fig:relocation}, and cross-image composition in Figure~\ref{fig:cross}. All results are generated in one round.

\begin{figure*}[!h]
    \centering
    \includegraphics[width=0.95\textwidth]{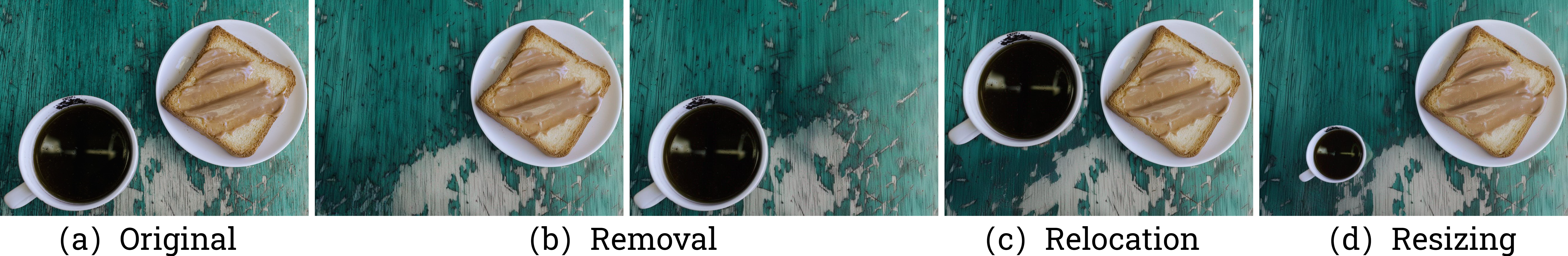}
    \caption{\textbf{Qualitative results of photorealistic image editing}.  We conduct basic editing operations to demonstrate our general editing ability, which is not limited to design images.}
    \label{fig:photo}
\end{figure*}

\vspace{2mm}
\noindent\textbf{Photorealistic Image Editing}
Section~\ref{subsec:cmp} and Figure~\ref{fig:magicbrush} shows the removal ability on the photorealistic dataset \textsc{MagicBrush}. Here we provide more qualitative results of different editing operations in Fig~\ref{fig:photo}.

\begin{figure*}[!h]
    \centering
    \includegraphics[width=0.95\textwidth]{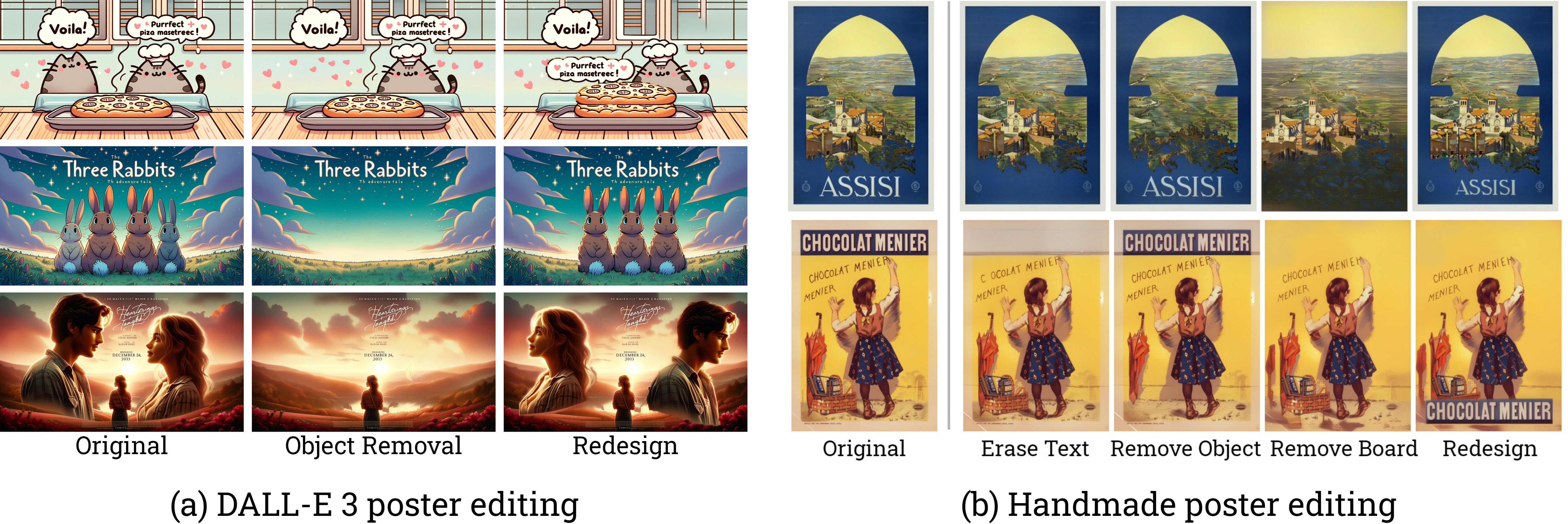}
    \caption{\textbf{Qualitative results of photorealistic image editing}. We show the results of removal and redesign on \dalle posters int (a) and handmade posters in (b).}
    \label{fig:poster}
\end{figure*}

\begin{figure*}[!h]
    \centering
    \includegraphics[width=0.92\textwidth]{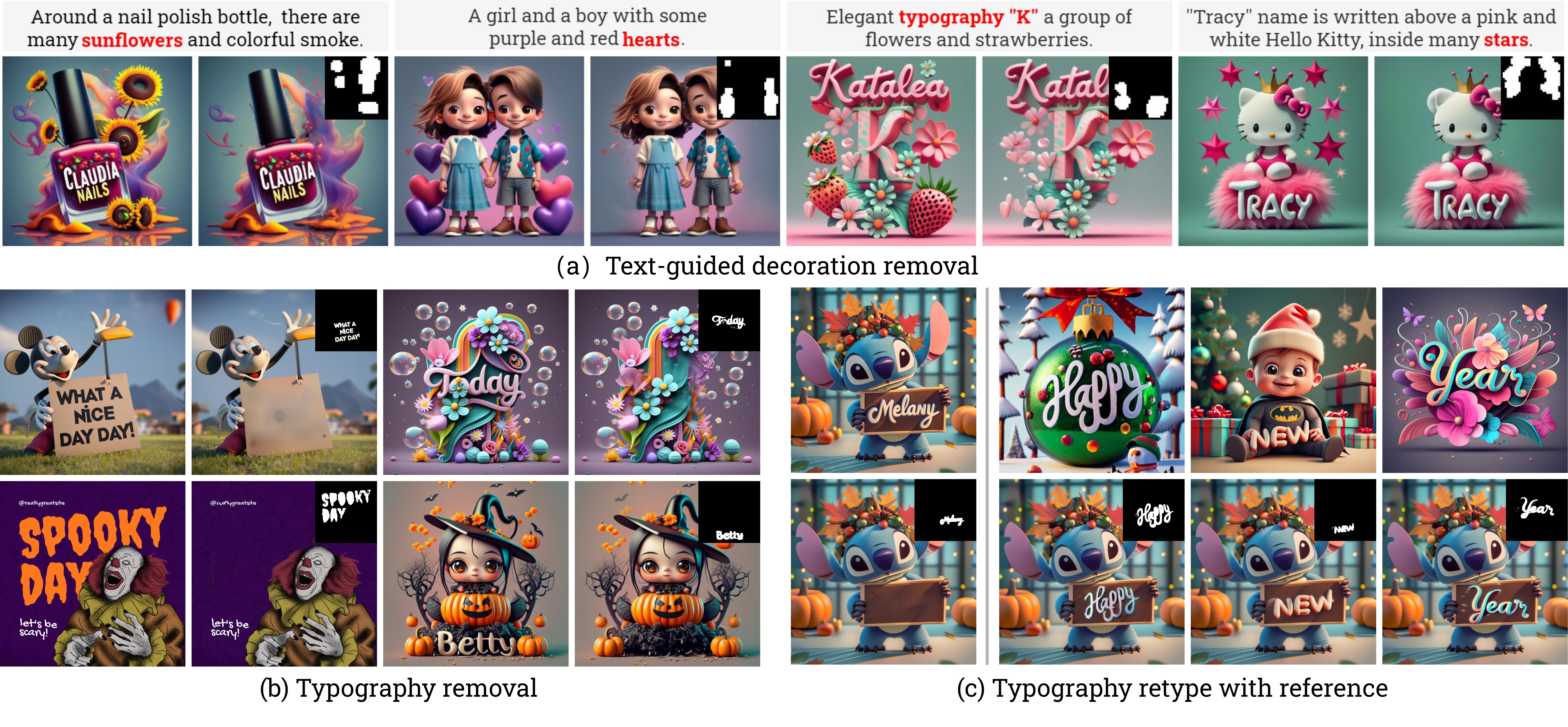}
    \vspace{-4mm}
    \caption{\textbf{Qualitative results of applications on design images.} (a) shows the results of decoration removal using the Cross-Attention mask, with the relevant token marked in \textcolor{red}{red}. (b) and (c) demonstrate the results of typography editing.}
    \label{fig:decoration}
\end{figure*}

\begin{figure*}[!h]
    \centering
    \includegraphics[width=0.92\linewidth]{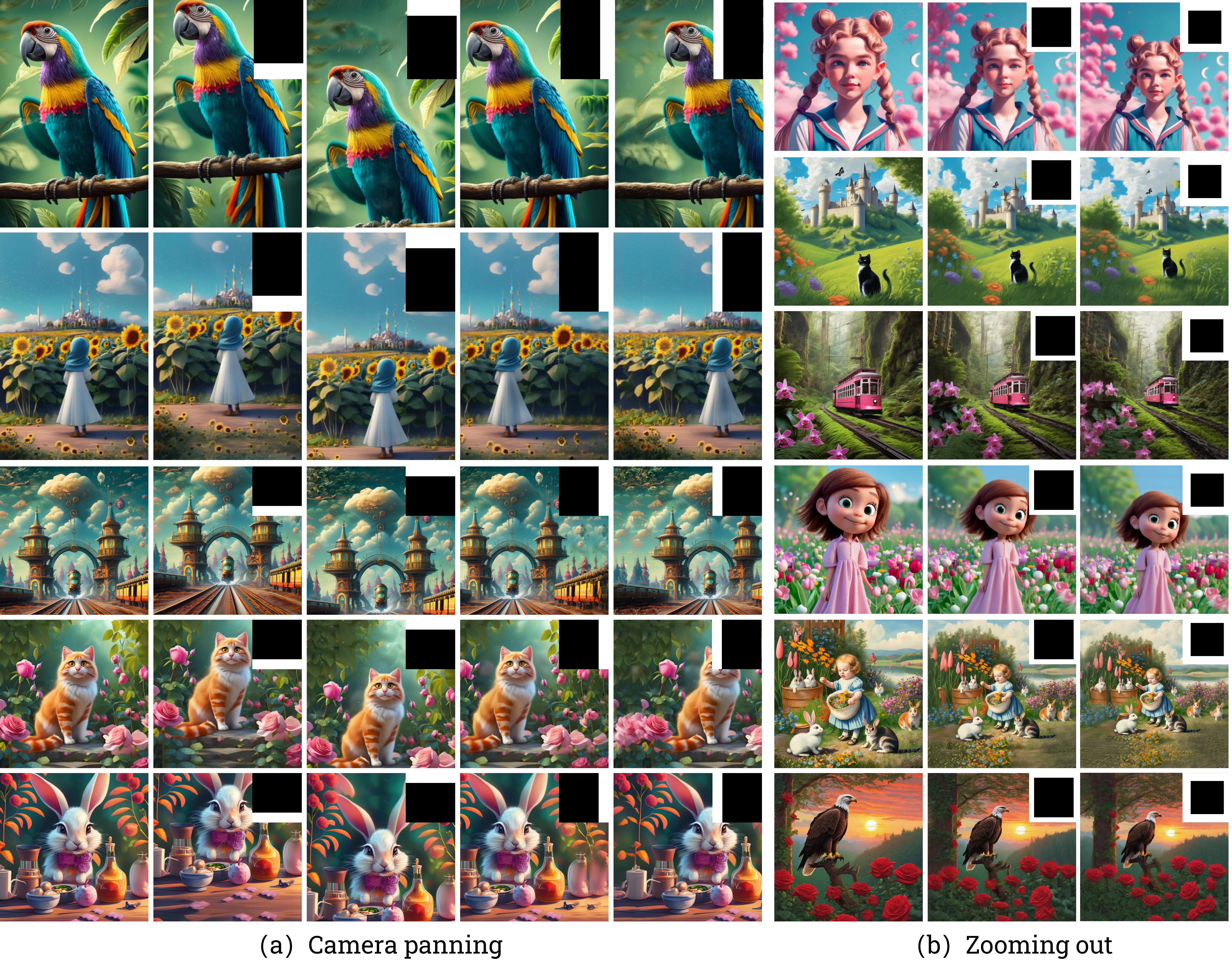}
    \vspace{-2mm}
    \caption{\textbf{Qualitative results of camera panning and zooming out tasks.} (a) presents the qualitative results of camera panning in four directions: up, down, left, and right, with a scale of 0.2 × H or 0.2 × W. (b) shows zooming out results at $1.25$ and $1.5$ scales.}
    \label{fig:pan+zoom}
\end{figure*}

\vspace{2mm}
\noindent\textbf{Applications on Design Images} Figure~\ref{fig:decoration} (b) illustrates the task of text-guided decoration removal with Cross-Attention masks, which are too irregular and numerous to mask manually. Figure~\ref{fig:decoration} (b) and (c) show the applications in typography editing on design images with object removal and cross-image composition. Figure~\ref{fig:poster} shows poster editing results. Figure~\ref{fig:pan+zoom} shows the camera panning and zooming out results on design images.

\vspace{-2mm}

%% file: sec/5_conclusion.tex
\section{Conclusion}
\label{sec:conclusion}
In this study, we propose a multi-layered latent decomposition and fusion framework that unifies various spatial-aware image editing operations without requiring additional tuning. To enhance image editing precision, we introduce two innovative techniques: a key-masking self-attention scheme and an artifact suppression scheme, aimed at improving the quality of background image layers and occluded object layers. Additionally, we utilize the layout planning capability of the advanced GPT-4V models to further refine our approach. Finally, we empirically validate the superiority of our method across a range of image editing tasks, particularly in the challenging domain of design images, through comprehensive quantitative and qualitative comparisons.